%% file: root.tex
\documentclass{article}
\usepackage{amssymb}
\usepackage[table]{xcolor}

\usepackage[final]{corl_2025} 

\usepackage[pdftex]{graphicx}
\usepackage{rpm_SIunits}
\usepackage{./rpm_acronyms}
\usepackage{rpm_math}
\usepackage{rpm_misc}

\usepackage{amsmath}
\usepackage{amssymb}  
\usepackage{subfigure}
\usepackage{subcaption}
\usepackage{multirow}
\usepackage{array,booktabs}
\usepackage{diagbox}
\usepackage{balance}
\usepackage{xspace}
\usepackage{algorithm}
\usepackage{algpseudocode}
\usepackage{adjustbox}
\usepackage{romannum}

\usepackage{cleveref}
\crefname{figure}{Fig.}{Figs.}
\Crefname{figure}{Fig.}{Figs.}
\usepackage{bbm}

\usepackage{textcomp}

\usepackage{soul,color}
\usepackage{lipsum}
\usepackage{dsfont}

\usepackage{xcolor}

\definecolor{imlprcolor}{HTML}{f3f7fc} 
\definecolor{bestcolor}{HTML}{def3e6} 
\definecolor{secondbestcolor}{HTML}{ecf8f1} 

\usepackage{amssymb}
\usepackage{pifont}
\newcommand{\cmark}{\ding{51}}%
\newcommand{\xmark}{\ding{55}}%
\usepackage{tikz} 
\usepackage{subcaption}
\usepackage{caption}
\title{\LARGE \bf ImLPR: Image-based LiDAR Place Recognition using Vision Foundation Models
}

\author{
  Minwoo Jung\\
  Department of Mechanical Engineering\\
  Seoul National University, 
  South Korea\\
  \texttt{moonshot@snu.ac.kr} \\
  \And
  Lanke Frank Tarimo Fu \\
  Department of Engineering Science \\
  University of Oxford,
  United Kingdom \\
  \texttt{fu@robots.ox.ac.uk} \\
  \AND
  Maurice Fallon \\
  Department of Engineering Science \\
  University of Oxford,
  United Kingdom \\
  \texttt{mfallon@robots.ox.ac.uk} \\
  \And
  Ayoung Kim${}^*$\\
  Department of Mechanical Engineering\\
  Seoul National University, 
  South Korea\\
  \texttt{ayoungk@snu.ac.kr} \\
}

\begin{document}
\maketitle

\begin{abstract}
    \ac{LPR} is a key component in robotic localization, enabling robots to align current scans with prior maps of their environment. While \ac{VPR} has embraced \acfp{VFM} to enhance descriptor robustness, LPR has relied on task-specific models with limited use of pre-trained foundation-level knowledge. This is due to the lack of 3D foundation models and the challenges of using \ac{VFM} with LiDAR point clouds. To tackle this, we introduce ImLPR, a novel pipeline that employs a pre-trained DINOv2 \ac{VFM} to generate rich descriptors for LPR. To the best of our knowledge, ImLPR is the first method to utilize a \ac{VFM} for \ac{LPR} while retaining the majority of pre-trained knowledge. ImLPR converts raw point clouds into novel three-channel \ac{RIV} to leverage \ac{VFM} in the LiDAR domain. It employs MultiConv adapters and Patch-InfoNCE loss for effective feature learning. We validate ImLPR on public datasets and outperform \ac{SOTA} methods across multiple evaluation metrics in both intra- and inter-session \ac{LPR}. Comprehensive ablations on key design choices such as channel composition, \ac{RIV}, adapters, and the patch-level loss quantify each component’s impact. We release ImLPR as \href{https://github.com/minwoo0611/ImLPR}{open source} for the robotics community. 
\end{abstract}

\keywords{LiDAR Place Recognition, Deep Learning, Vision Foundation Model} 



\input{contents/1_introduction}

\input{contents/2_relatedwork}
\input{contents/3_method}

\input{contents/4_experiment}
\input{contents/5_conclusion}
\input{contents/6_limitation}

\section*{ACKNOWLEDGMENT}
This project has been partly funded by the National Research Foundation of Korea (NRF) grant funded by the Korea government (MSIT) (No. RS-2024-00461409), the UKRI project Mobile Robotic Inspector (EP/Z531212/1), and a Royal Society Univ. Research Fellowship (Fallon). 

\balance
\bibliography{references}

\newpage
\appendix
\section*{Appendices}
\input{supplementary/supplementary}
\end{document}

%% file: contents/1_introduction.tex
\section{Introduction}
\label{sec:intro}

Place recognition, encompassing both \ac{VPR} and \ac{LPR}, provides an initial localization estimate when seeking to determine if a location has been previously visited, using a prior map of the location. Early learning-based methods relied on specialized architectures trained on domain-specific datasets~\cite{arandjelovic2016netvlad, hausler2021patch, vidanapathirana2022logg3d, komorowski2022improving, xia2023casspr, luo2024bevplace++}. While \ac{VPR} has advanced by using VFMs~\cite{garg2024revisit, izquierdo2024optimal, keetha2023anyloc, nie2024mixvpr++}, pre-trained on large-scale datasets~\cite{kirillov2023segment, oquab2023dinov2, radford2021learning}, \ac{LPR} has been limited by the lack of suitable 3D foundation models and the modality gap when adapting VFMs to LiDAR data.

\begin{figure}[t]
    \centering
    \includegraphics[width=.96\columnwidth]{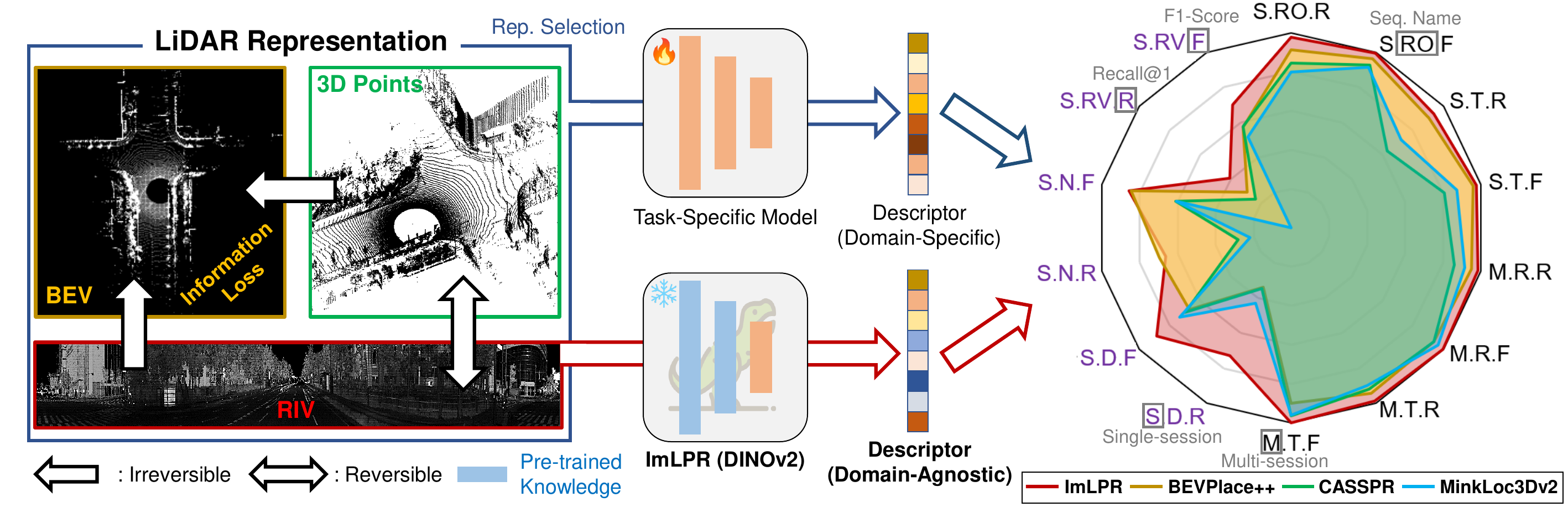}
    \vspace{-1mm}
    \caption{Without using a foundation model, traditional \ac{LPR} relies on domain-specific training with 3D point clouds, BEV, or RIV. To leverage a \ac{VFM}, ImLPR uses RIV images to capture geometric information while avoiding the information loss that the BEV projection induces. The radar chart (right) highlights ImLPR’s superior performance in trained (black) and unseen (purple) domains.}
    \vspace{-1mm}
    \label{fig:overall_diagram}
    \vspace{-4mm}
\end{figure}

Efforts applying foundation models on LiDAR data can be characterized in two ways: developing full 3D foundation models or converting LiDAR data into a format that can be used with \ac{VFM}. Existing 3D foundation models~\cite{wu2024point, wu2025sonata}, designed for object detection or indoor scene analysis, are unsuitable for \ac{LPR} due to their focus on specific tasks and the heavy computation required when working with 3D data with large sets of parameters. On the other hand, converting 3D point clouds into 2D images is a more straightforward approach for leveraging \ac{VFM} for \ac{LPR}. 
Two types of image projection are typically used to create a 2D representation from a 3D point cloud, \ac{BEV} and \ac{RIV}, as shown in \figref{fig:overall_diagram}.
However, a substantial domain gap exists between natural images used to pre-trained VFMs and projected LiDAR images, particularly when using single-channel \cite{luo2024bevplace++} or overly large multi-channel inputs \cite{chen2020rss}. Previous approaches that rely on such representations struggle to effectively leverage pre-trained three-channel vision models.

To tackle these issues, we present ImLPR, a novel pipeline for \ac{LPR} that adapts DINOv2~\cite{oquab2023dinov2}, a \ac{SOTA} VFM with robust and transferable feature extraction capabilities. 
We have designed our system to around a RIV which encodes point clouds using reflectivity, range, and normal ratio. This allows RGB-pretrained VFM to extract discriminative LiDAR features
Our empirical results show that each proposed channel has a significant impact on \ac{LPR} performance, highlighting the importance of channel design.
Notably, the three-channel configuration proves more effective for RIV than for BEV.
Furthermore, by inserting and training lightweight adapters, we preserve most of the pre-trained DINOv2 weights while adapting it effectively for LPR using RIV input images. Additionally, we propose the Patch-InfoNCE loss, a patch-level contrastive loss, enhancing discriminability and robustness by ensuring consistent feature representations across corresponding RIV patches. Combining LiDAR’s geometric precision and semantic information with DINOv2’s representational power, ImLPR surpasses \ac{SOTA} \ac{LPR} methods. Our contributions include:

\begin{itemize}
    \item ImLPR is the first LPR pipeline using a VFM while retaining the majority of pre-trained knowledge. Our key innovation lies in a tailored three-channel RIV representation and lightweight convolutional adapters, which seamlessly bridge the 3D LiDAR and 2D vision domain gap.
    Freezing most DINOv2 layers preserves pre-trained knowledge during training, ensuring strong generalization and outperforming task-specific LPR networks.
    \item We introduce the Patch-InfoNCE loss, a patch-level contrastive loss, to enhance the local discriminability and robustness of learned LiDAR features. We demonstrate that our patch-level contrastive learning strategy achieves a performance boost in \ac{LPR}.
    \item ImLPR demonstrates versatility on multiple public datasets, outperforming \ac{SOTA} methods. Furthermore, we also validate the importance of each component of the ImLPR pipeline.
\end{itemize}

%% file: contents/2_relatedwork.tex
\section{Related Work}
\label{sec:relatedwork}

\subsection{Traditional LiDAR Place Recognition}
Numerous works have studied \ac{LPR}. Early methods like PointNetVLAD~\cite{uy2018pointnetvlad} used PointNet for feature extraction, while others~\cite{liu2019lpd, guo2019local} refined descriptors with MLPs. Efficiency-driven approaches, such as OverlapTransformer~\cite{ma2022overlaptransformer} and CVTNet~\cite{ma2023cvtnet}, projected point clouds into range images for 2D convolutions. Sparse convolution methods~\cite{vidanapathirana2022logg3d, komorowski2022improving} optimized 3D processing, while BEV projections~\cite{luo2023bevplace} have been used to capture the global contour of a scene. Recent efforts, like SE(3)-equivariant networks~\cite{lin2023se}, leverage rotation-invariant features, and SeqMatchNet~\cite{garg2022seqmatchnet} uses contrastive learning with sequence matching for robustness. However, these methods, trained on specific datasets, struggle to adapt to scene variations and exhibit reduced performance when tested in different domains. Their domain-specific descriptors struggle with diverse outdoor conditions, hindering broad applicability. Inspired by the progress of \ac{VPR}~\cite{garg2024revisit, izquierdo2024optimal}, we address these limitations by introducing VFMs to the task of \ac{LPR}, leveraging the rich, robust descriptors of foundation models to overcome the traditional domain-specific constraints described above.

\subsection{Vision and 3D Foundation Models for 3D Data}
VFMs~\cite{oquab2023dinov2, radford2021learning, kirillov2023segment} excel in \ac{VPR} due to their robust feature representations~\cite{garg2024revisit, keetha2023anyloc}. In particular, DINOv2 was pre-trained on massive datasets containing 140 million images. In contrast, 3D foundation models like PTv3~\cite{wu2024point} and Sonata~\cite{wu2025sonata} are trained on smaller datasets with 23-139 thousand point clouds, and focus on object-level tasks, limiting their generality for outdoor \ac{LPR}.
Recently, researchers have attempted to apply \ac{VFM} to point clouds \cite{puy2024three, vodisch2025lidar}. However, they focus on object detection or registration, and still rely on visual images, diverging from \ac{LPR}. 
Instead, to adapt LiDAR data for use with image-based networks, traditional methods project point clouds into 2D formats like \ac{BEV}~\cite{luo2023bevplace} or \ac{RIV}~\cite{chen2020rss}, and then use these images with ImageNet-pretrained CNNs such as ResNet~\cite{he2016deep}. However, compared to \ac{VFM}, these methods lack the expressiveness needed for diverse outdoor scenes and are limited by their small training datasets and simple architectures. Additionally, domain differences between LiDAR images and visual images hinder the direct application of VFMs. 
Fine-tuning the entire model such as LIP-Loc~\cite{Shubodh_2024_WACV}, can lead to catastrophic forgetting, reducing generalization by forgetting pre-trained knowledge.
To address these challenges, we utilize DINOv2 with a MultiConv adapter to bridge domain differences while preserving almost the entire pre-trained knowledge within the \ac{VFM} network. We adopt the \ac{RIV} projection, which captures denser geometric detail and is more similar to normal visual images compared to the sparser BEV images~\cite{luo2023bevplace, luo2024bevplace++}.

\begin{figure*}[!t]
    \centering
    \includegraphics[width=.96\textwidth]{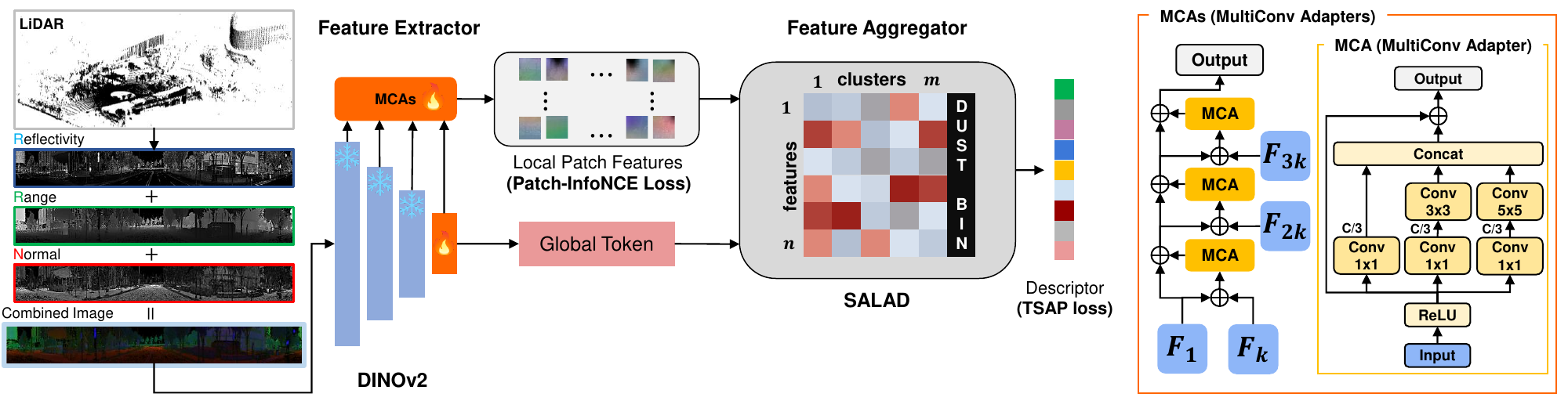}
    \caption{The point cloud is projected into a RIV image containing reflectivity, range, and normal channels. A pre-trained DINOv2 model, adapted via MultiConv adapters (MCAs), extracts rich patch-level features. Patch-InfoNCE loss enhances local feature discriminability, while SALAD aggregates features into a global descriptor. This effectively leverages the \ac{VFM} for use with LiDAR.}
    \vspace{-1mm}
    \label{fig:pipeline}
    \vspace{-4.5mm}
\end{figure*}

%% file: contents/3_method.tex
\section{Methodology}
\subsection{Overview of ImLPR}
ImLPR first renders the scan as an RGB-style image $\mathbf{I} \in \mathbb{R}^{H \times W \times 3}$ and then computes a global representation $\mathbf{g} \in \mathbb{R}^{t}$ through the mapping function $\Omega = h \circ f$. The \ac{VFM}-based encoder $f: \mathbb{R}^{H \times W \times 3} \rightarrow \mathbb{R}^{H'W' \times C}$ ($H' = \lfloor H/14 \rfloor,\; W' = \lfloor W/14 \rfloor$) extracts patch features, which are then pooled into the descriptor $\mathbf{g}$ by the aggregator $h: \mathbb{R}^{H'W' \times C} \rightarrow \mathbb{R}^{t}$. Metric learning optimizes $\Omega$ such that, if the spatial distance $\mathcal{D}(q, i) < \mathcal{D}(q, j)$, then the descriptor distance $d_g(q, i) < d_g(q, j)$, enforcing this relationship at both the feature and global descriptor levels.

\subsection{Input Data Processing}
To convert raw LiDAR point clouds into structured images, we map each point \(p_i = (x_i, y_i, z_i) \in \mathbf{P}\) to RIV pixel coordinates \((u_i, v_i)\) via:
\begin{equation}
\small
\begin{pmatrix}
u_i \\[6pt]
v_i
\end{pmatrix}
=
\begin{pmatrix} 
0.5 \left[\,1 - \arctan(y_i, x_i) / {\pi}\right] \cdot W \\[8pt] 
\left[\,1 - (\arcsin(z_i/r_i) + f_{\text{up}}) / {f}\right] \cdot H
\end{pmatrix},
\normalsize
\end{equation}
where \(W\) and \(H\) denote the image width and height, \(r_i\) is the range from the origin to \(p_i\), \(f_{\text{up}}\) is the upper bound of the vertical \ac{FOV}, and \(f\) denotes the total vertical \ac{FOV}. 
To further enrich the image with geometric information, we include the three LiDAR channels—reflectivity, range, and normal ratio—in each pixel. Reflectivity offers semantic distinction between objects with different surface properties, while range encodes geometric features of the scene, such as object distance and depth variation. 
The normal ratio is derived from the singular value ratio of the covariance matrix for the $k$-nearest neighboring points of $p_i$. Singular Value Decomposition (SVD) is applied to this covariance matrix, and the logarithmic ratio of the largest to smallest singular values encodes the normal information. This scalar effectively summarizes local geometric variations—such as surface planarity and edge-like structures—into a single channel, offering a computationally efficient alternative to multi-channel normal vectors \cite{chen2020rss}.
This multi-channel image is used as input for ImLPR, as depicted in \figref{fig:pipeline}.

\subsection{Feature Extraction with Adapted DINO ViT}
We adopt a DINO ViT-S/14 model pre-trained on RGB images and tailor it to LiDAR RIV images by fine-tuning only the final two transformer blocks of the model. Similar to SelaVPR++ \cite{lu2025selavpr++}, we bridge the gap between the DINO pre-training RGB dataset and new LiDAR data by postprocessing the per-patch DINO features using lightweight MultiConv adapter layers. Concretely, we insert these adapters at regular intervals among the frozen layers, allowing them to refine intermediate patch representations while keeping most of the network frozen. This design preserves the rich pre-trained capabilities of the model while facilitating LiDAR-specific adaptation with lower computational overhead compared to full end-to-end fine-tuning.
Since MultiConv adapters are placed at \(k\) regular intervals, this further optimizes memory and computation. Formally, let \(x_l\) represent the intermediate DINO feature of the \(l\)-th transformer block, comprising both patch features \(x_l^\text{patch}\) and token features \(x_l^\text{token}\). We define the adapter-refined patch features \(y_i\) as:
\begin{equation}
y_i =
\begin{cases}
\small
\text{Adapter}\bigl(x_{i}^\text{patch} + x_{ik}^\text{patch}\bigr) + x_{i}^\text{patch}, & i = 1 \\[6pt]
\small
\text{Adapter}\bigl(y_{i-1}^\text{patch} + x_{ik}^\text{patch}\bigr) + y_{i-1}^\text{patch}, & 2 \leq i \leq \lfloor L / k \rfloor
\end{cases}
\normalsize
\end{equation}
where \(L\) is the total number of transformer blocks. The token features \(x_l^\text{token}\) remain untouched. By focusing on the refined patch features, our method efficiently leverages strong generalizable visual representations from the pre-trained model while accommodating LiDAR-specific structures.

\subsection{Feature Aggregation with Optimal Transport}
We aggregate the refined patch features into a global representation using an optimal transport approach similar to SALAD \cite{izquierdo2024optimal}. Let \(\mathbf{F} \in \mathbb{R}^{H'W' \times C}\) be the local features derived from the adapter-refined patch features. A convolutional layer first maps \(\mathbf{F}\) to a score matrix \(\mathbf{S} \in \mathbb{R}^{n \times m}\), representing the assignment probabilities of local features to \(m\) learnable cluster centers. We then apply optimal transport, specifically the Sinkhorn algorithm \cite{NIPS2013_af21d0c9}, to obtain an optimized assignment matrix \(\mathbf{R} \in \mathbb{R}^{n \times m}\) by iteratively normalizing the rows and columns of the exponentiated score matrix. The aggregated feature \(\mathbf{V} \in \mathbb{R}^{m \times l}\) is computed as $V_{j,k} = \sum_{i=1}^{n} R_{i,j}\,\bar{F}_{i,k},
$ where \(\bar{\mathbf{F}}\) denotes the intermediate feature embeddings obtained by applying a convolution layer to the original feature \(\mathbf{F}\).

To capture global structural context, we generate a compact global embedding \(\mathbf{G} \in \mathbb{R}^{e}\) via linear layers. The global token passes through two linear layers to compress and capture global context. The final descriptor combines flattened local features and the global embedding as \(g = [\mathbf{V}.flatten(), \mathbf{G}] \in \mathbb{R}^{m \times l + e}\). This fusion of clustering-based local features and global embedding captures both detailed local patterns and robust global structure.

\subsection{Patch-level Contrastive Learning}
To encourage learning both locally (patch feature-level) and globally (descriptor-level), we introduce a patch-level contrastive loss in addition to the global objective. This promotes effective gradient flow throughout the network. For patch-level supervision, we mine positive and negative patch pairs from two RIV range images based on geometric consistency.

\begin{figure}[!t]
    \centering
    \includegraphics[width=.85\columnwidth]{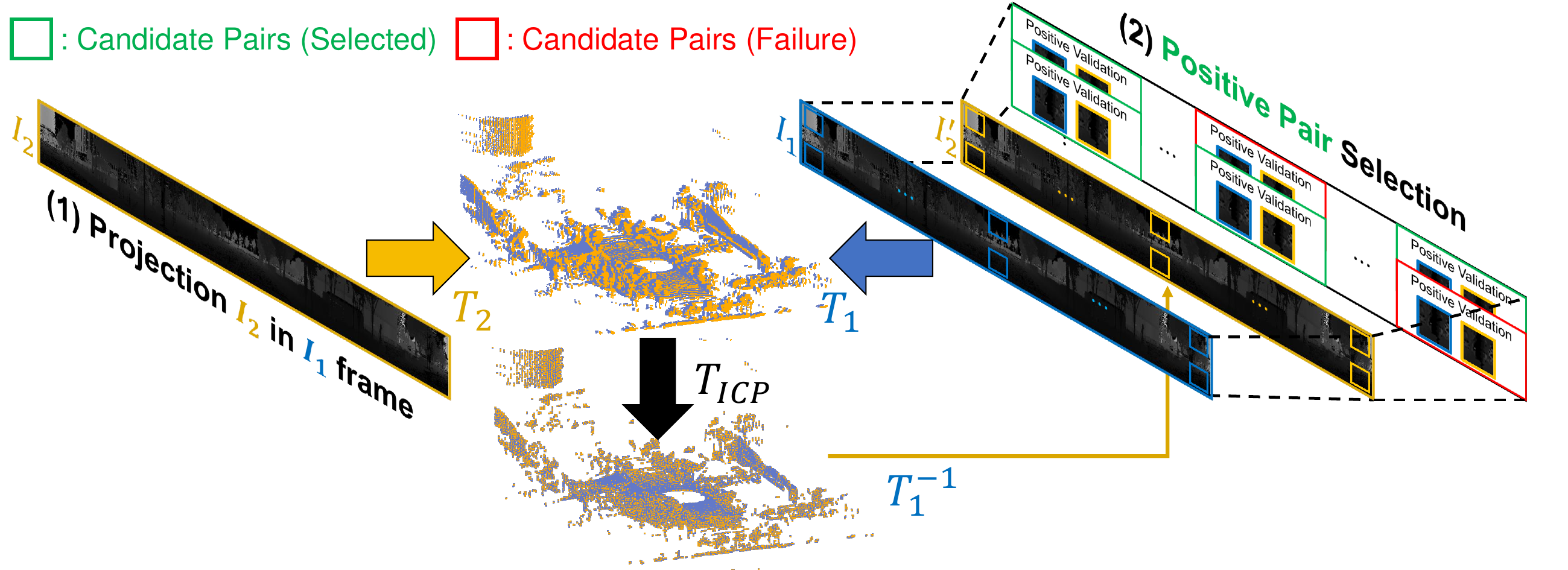}
    \vspace{-0.7mm}
    \caption{Patch correspondence pipeline aligns two RIV images by transforming their point clouds to a shared coordinate system using known poses and Iterative Closest Point (ICP). Positive patch pairs are validated based on spatial overlap and range similarity, while negatives ensure large spatial separation. This process supplies pairs for Patch-InfoNCE loss to train discriminative local features.}
    \label{fig:correspondence}
    \vspace{-5mm}
\end{figure}

\textbf{Positive Patch Mining: }We transform both range images into 3D point clouds using their RIV geometry and align them in a common world frame using ground truth poses refined by ICP. Points from the second scan are reprojected onto the first RIV to identify candidate patch correspondences, as illustrated in \figref{fig:correspondence}. Positive pairs are selected based on pixel overlap and range similarity. A patch pair \((p_1, p_2)\) is considered positive if the ratio of valid overlapping pixels \(\mathcal{P}_{\text{ov}}\), containing range values in both patches, exceeds \(\rho_{\text{valid}} = 0.5\). For all pixels in \(\mathcal{P}_{\text{ov}}\), we compute the per-pixel range differences $d_i = r_{1,i} - r_{2,i}$. An adaptive threshold $\tau_r$ is computed using the median absolute deviation (MAD): $\tau_r = \text{median}(d_i) + k \cdot \text{MAD}$, ensuring robustness to measurement noise. A pair is confirmed positive if the mean range difference $\Delta_r = \frac{1}{|\mathcal{P}_{\text{ov}}|} \sum_{i \in \mathcal{P}_{\text{ov}}} |r_{1,i} - r_{2,i}|$ is less than $\tau_r$. This ensures that spatially consistent, range-coherent positive patch pairs are selected.

\textbf{Negative Patch Mining: }To select negative patch pairs, we enforce large vertical and horizontal distances (\(v_{\text{dist}}\) and \(h_{\text{dist}}\)) from the positive patches to avoid spatial overlap. The cylindrical geometry is considered, ensuring that patches near the \(0^\circ\) and \(360^\circ\) boundaries are treated as spatially distant.

To supervise patch-level contrastive learning, we propose Patch-InfoNCE, an extension of the InfoNCE loss~\cite{radford2021learning} for local features. Unlike global descriptor losses, Patch-InfoNCE operates on local embeddings from features \(\mathbf{F_1}, \mathbf{F_2} \in \mathbb{R}^{H'W' \times C}\) of two adjacent \ac{RIV} images. It supports multiple positive patch pairs \((p_1, p_2)\) per image and computes their cosine similarity. For each positive pair \(F_1^{p_1}\) and \(F_2^{p_2}\), we select hard negatives: \(n_1\) from \(\mathbf{F_1}\) relative to \(p_2\), and \(n_2\) from \(\mathbf{F_2}\) relative to \(p_1\). The Patch-InfoNCE loss is defined as:
\begin{equation}
\small
\mathcal{L}_{P} = -\frac{1}{|\mathcal{P}|}\sum_{(p_1,p_2)\in\mathcal{P}}\log\!\Biggl(\frac{\exp(s_{\text{p}}/\tau_l)}{\exp(s_{\text{p}}/\tau_l) + \sum_j \exp(s_{\text{n},j}/\tau_l)}\Biggr),
\normalsize
\end{equation}
where \(\mathcal{P}\) is the set of positive pairs, \(\tau_l\) is the temperature, and \(s_{\text{p}}\), \(s_{\text{n},j}\) denote the cosine similarities of positive and negative pairs, respectively. This loss encourages consistent local features at corresponding spatial locations across \ac{RIV} images, enabling discriminative patch-level embeddings for \ac{LPR}. The final objective combines both local and global terms as \(\mathcal{L}_{\text{final}} = \mathcal{L}_{P} + \lambda \mathcal{L}_{\text{TSAP}}\), where \(\lambda\) balances their ratio. Details of \(\mathcal{L}_{\text{TSAP}}\) \cite{komorowski2022improving} are provided in the appendix. This dual-loss formulation captures both fine-grained context and global semantics, improving overall \ac{LPR} performance.

%% file: contents/4_experiment.tex
\section{Experiment}
\label{sec:experiment}
\subsection{Experimental Setup}

\textbf{Implementation Details:} ImLPR was implemented in PyTorch using a DINO ViT-S/14, training only the last two transformer blocks and inserting MultiConv adapters every three blocks. The size of the RIV image used is \(H \times W = 126 \times 1022\). Additional details are available in the appendices.

\textbf{Datasets \& Evaluation Metrics:} ImLPR is evaluated on HeLiPR (\texttt{O}: OS2-128 and \texttt{V}: VLP-16C)~\cite{jung2024helipr}, MulRan (OS1-64)~\cite{kim2020mulran}, and NCLT (HDL-32E)~\cite{carlevaris2016university} datasets, with scans sampled at 3\unit{m} intervals. 
From HeLiPR, we use sequences \texttt{Roundabout01-03} and \texttt{Town01-03}, where the suffix \texttt{O} or \texttt{V} indicates the LiDAR used. From MulRan, we use the \texttt{DCC01-03} sequences for evaluation. 
We use nine HeLiPR sequences with the Ouster for training: \texttt{DCC04-06}, \texttt{KAIST04-06}, and \texttt{Riverside04-06}.
To handle varying point cloud density, 3D sparse convolution methods downsample to 8192 points, while other parameters retain default settings.
ImLPR is compared against LoGG3D-Net~\cite{vidanapathirana2022logg3d}, MinkLoc3Dv2~\cite{komorowski2022improving}, CASSPR~\cite{xia2023casspr}, and BEVPlace++~\cite{luo2024bevplace++}, using Recall@1 (R@1) and maximum F1 score. 
All methods are trained under the same conditions. Positive pairs are defined as matches within \unit{10}{\meter}, assuming a four-lane highway. During training, negative pairs are sampled beyond \unit{30}{\meter}, while during evaluation, pairs outside the \unit{10}{\meter} range are treated as negatives.

\input{tab/intra_results}

\begin{figure}[!t]
    \centering
    \includegraphics[trim = 0cm 0cm 0cm 0cm, width=0.85\columnwidth]{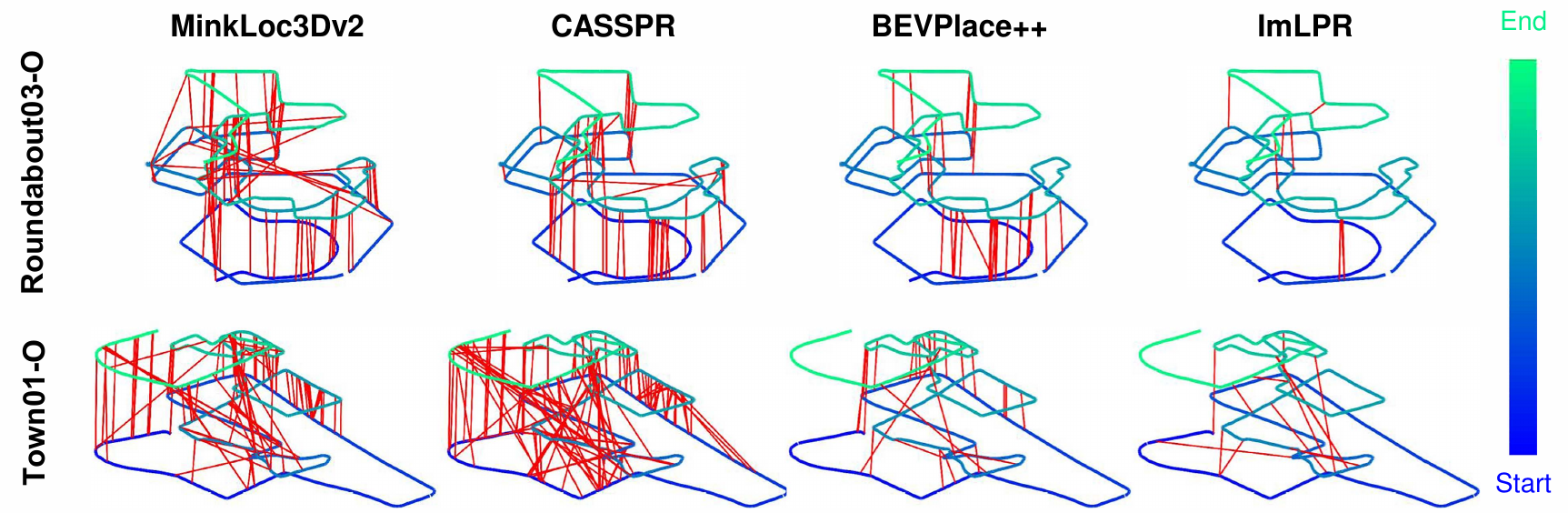}
    \vspace{-1mm}
    \caption{Trajectory from two sequences, with red lines marking false positives (fewer red lines indicate better performance). It highlights ImLPR having fewer errors compared to \ac{SOTA} methods.}
    \label{fig:intrasession}
    \vspace{-6mm}
\end{figure}

\subsection{Intra-session Place Recognition}
Scans within 60 seconds of the query are excluded to prevent intra-session matching (i.e., matching within the same trajectory), and processing begins only after 90 seconds to ensure a sufficient database. As shown in \tabref{tab:intra} and \figref{fig:intrasession}, ImLPR surpasses all baselines on \texttt{Roundabout-O} and \texttt{Town-O}. LoGG3D-Net exhibits lower recall due to the domain shift between training and testing. BEVPlace++ ranks second, capturing scan contours but underperforms ImLPR in narrow alleys and small-scale environments, as ImLPR leverages RIV representations for fine-grained feature extraction. 3D sparse convolution methods like MinkLoc3Dv2 and CASSPR rank third, limited by the absence of large-scale pre-trained models. ImLPR achieves the highest R@1 and F1 score, demonstrating robust and discriminative intra-session \ac{LPR} performance.

\subsection{Inter-session Place Recognition}
\begin{figure}[!t]
    \centering
    \includegraphics[trim = 0cm 0cm 0cm 0cm, width=.94\columnwidth]{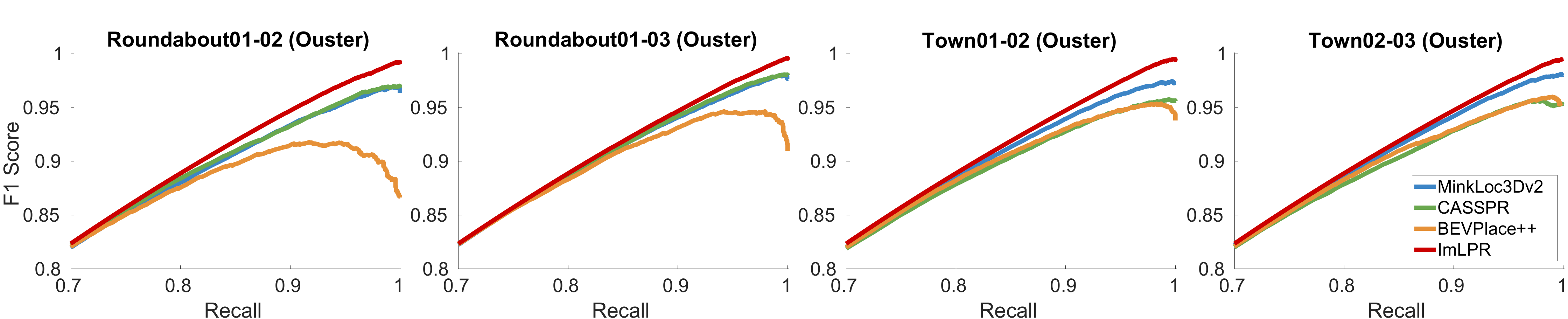}
    \vspace{-2mm}
    \caption{F1-Recall curve for inter-session place recognition. ImLPR consistently outperforms all baselines across all sequence pairs. BEVPlace++ exhibits a low F1 hindered by the projection-induced information loss. This highlights ImLPR’s superior balance of precision and recall.}
    \label{fig:intersession}
    \vspace{-6.5mm}
\end{figure}

\tabref{tab:inter_roundabout} and \tabref{tab:inter_town} show inter-session results (i.e., between scans collected at different times) for \texttt{Roundabout-O} and \texttt{Town-O}. Each result is shown in the format \texttt{Database-Query}. ImLPR outperforms other methods across both environments, adapting to session-induced variations. BEVPlace++ achieves the second-highest Recall@1, but its lower F1 score indicates reduced precision. We believe this is due to the simple network consisting of 2D convolution without additional mechanisms such as attention. Furthermore, its BEV representation, projecting all data into a single plane, limits its ability to distinguish subtle session-specific variations. As shown in \figref{fig:intersession}, BEVPlace++ has a lower F1-Recall curve than MinkLoc3Dv2 and CASSPR, despite higher recall. Since the F1-Recall curve captures the trade-off between precision and recall across thresholds, a lower curve suggests a higher false positive rate. MinkLoc3Dv2 and CASSPR follow ImLPR in F1 score with inconsistent results across sequences. This variability highlights the benefit of using large-scale pretrained models for consistent results. LoGG3D-Net ranks lowest due to domain mismatches. ImLPR’s RIV representations precisely capture fine-grained details as shown in \figref{fig:dino_riv}, and its network design ensures top inter-session \ac{LPR} performance.

\input{tab/inter_roundabout}
\input{tab/inter_town}

\begin{figure}[H]
    \vspace{-9.5mm}
    \centering
    \begin{minipage}{0.4\textwidth} 
        \centering   
        \begin{table}[H]
            \caption{Generalization Assessment}
            \centering
            \resizebox{\linewidth}{!}{%
\begin{tabular}{l||cc|cc|cc|cc}
\toprule
\multirow{2}{*}{Method} & \multicolumn{2}{c|}{\texttt{DCC}} & \multicolumn{2}{c|}{\texttt{NCLT}} & \multicolumn{2}{c|}{\texttt{Roundabout-V}} & \multicolumn{2}{c}{Average} \\
 & AR@1 & AF1 & AR@1 & AF1 & AR@1 & AF1 & AR@1 & AF1 \\
\midrule
LoGG3D-Net & 0.094 & 0.214 & 0.149 & 0.434 & 0.131 & 0.394 & 0.125 & 0.347 \\
MinkLoc3Dv2 & \underline{0.722} & \underline{0.870} & 0.622 & 0.808 & 0.513 & 0.764 & 0.619 & 0.814 \\
CASSPR & 0.683 & 0.851 & 0.652 & 0.810 & 0.630 & \underline{0.793} & 0.655 & 0.818 \\
BEVPlace++ & 0.678 & 0.839 & \textbf{0.850} & \underline{0.922} & \underline{0.655} & 0.790 & \underline{0.728} & \underline{0.850} \\
ImLPR & \textbf{0.865} & \textbf{0.943} & \underline{0.834} & \textbf{0.927} & \textbf{0.712} & \textbf{0.852} & \textbf{0.804} & \textbf{0.907} \\
\bottomrule
\end{tabular}%

            }      
            \label{tab:sensor_dcc}
\tiny
\vspace{-0.1mm}
{$\,$}

\tiny 
\raggedright 
AR@1: Average Recall@1, AF1: Average F1 score
\vspace{-1mm}
        \end{table}

    \end{minipage}%
    \hfill
    \begin{minipage}{0.58\textwidth}
\begin{table}[H]
\centering
\caption{Performance Impact of RIV Image Channels}
\label{tab:ab_channel}
\resizebox{\textwidth}{!}{%
\begin{tabular}{l||ccc||cc|cc|cc|cc}
\toprule
\multirow{2}{*}{Method} & \multicolumn{3}{c||}{Image Channel} & \multicolumn{2}{c|}{\texttt{Roundabout-O}} & \multicolumn{2}{c|}{\texttt{Town-O}} & \multicolumn{2}{c|}{\texttt{DCC}} & \multicolumn{2}{c}{Average} \\
& CH1 & CH2 & CH3 & AR@1 & AF1 & AR@1 & AF1 & AR@1 & AF1 & AR@1 & AF1 \\
\midrule
\texttt{ExpA-1} & \ding{52} &  &  & 0.946 & 0.973 & 0.956 & 0.979 & 0.773 & 0.884 & 0.892 & 0.945 \\
\texttt{ExpA-2} & \ding{52} & \ding{52} &  & \underline{0.975} & \underline{0.989} & \underline{0.980} & \underline{0.990} & \underline{0.901} & \underline{0.951} & \underline{0.952} & \underline{0.977} \\
\texttt{ExpA-3} & \ding{52} & \ding{52} & \ding{52} & \textbf{0.979} & \textbf{0.990} & \textbf{0.985} & \textbf{0.993} & \textbf{0.945} & \textbf{0.970} & \textbf{0.970} & \textbf{0.984} \\
\bottomrule
\end{tabular} \label{tab:ab_channel_main}
} 
\tiny
\vspace{-0.1mm}
{$\,$}

\tiny 
\raggedright 
CH1: Reflectivity, CH2: Range, CH3: Normal Ratio
\vspace{-1mm}
\end{table}

    \end{minipage}
    \vspace{-5mm}
    \label{fig:dcc}
\end{figure}

\subsection{Ablation Studies}
\textbf{Model Generalization Assessment: }To assess the generality of ImLPR, we conducted intra-session \ac{LPR} on MulRan \texttt{DCC}, NCLT (\texttt{2012-01-08, 01-15, 01-22}), and HeLiPR \texttt{Roundabout-V} without additional training. As these datasets lack a reflectivity channel, normalized intensity was used. For NCLT and \texttt{Roundabout-V}, all methods were validated using accumulated scans as both query and database. The average performance across three sequences is reported in \tabref{tab:sensor_dcc}, with dataset-specific details provided in the appendix.
In \texttt{DCC}, while all methods experienced performance declines, ImLPR demonstrated the most robust performance, excelling in handling sensor differences and temporal shifts. In \texttt{NCLT} and \texttt{Roundabout-V}, it effectively generalized to unseen sensors and environments. Although discrete and noisy intensity from \texttt{NCLT} degraded the semantic quality of RIV images (\figref{fig:intensity_dist}), ImLPR still attained the highest F1 score by leveraging geometric cues. Unlike competing methods that showed inconsistent rankings across datasets—reflecting sensitivity to specific domains—ImLPR consistently maintained best performance as shown in \figref{fig:overall_diagram} and \tabref{tab:sensor_dcc}. This stability highlights its strong generalization, enabled by RIV and \ac{VFM} with adapter.

\textbf{Image Channels: }ImLPR uses three image channels—reflectivity, range, and normal ratio—to represent LiDAR point clouds. To assess their individual contributions, we evaluate three configurations: reflectivity only (\texttt{ExpA-1}), reflectivity and range (\texttt{ExpA-2}), and all three channels (\texttt{ExpA-3}), excluding Patch-InfoNCE loss since it relies on the range channel. We average inter-session place recognition results across all sequences, as shown in \tabref{tab:ab_channel_main}. Each added channel improves AR@1 and AF1. Reflectivity provides a strong semantic baseline, while range adds geometric depth and significantly boosts performance. The normal ratio further improves accuracy by capturing local 3D structure, though its effect is smaller due to the similarity with range, as both channels encode geometric information. These results demonstrate that each channel contributes distinct, complementary features that collectively strengthen the performance of ImLPR.

\begin{figure}[!t]
    \centering
    \subfigure[Feature Visualization with BEV]{
        \includegraphics[width=.3\columnwidth]{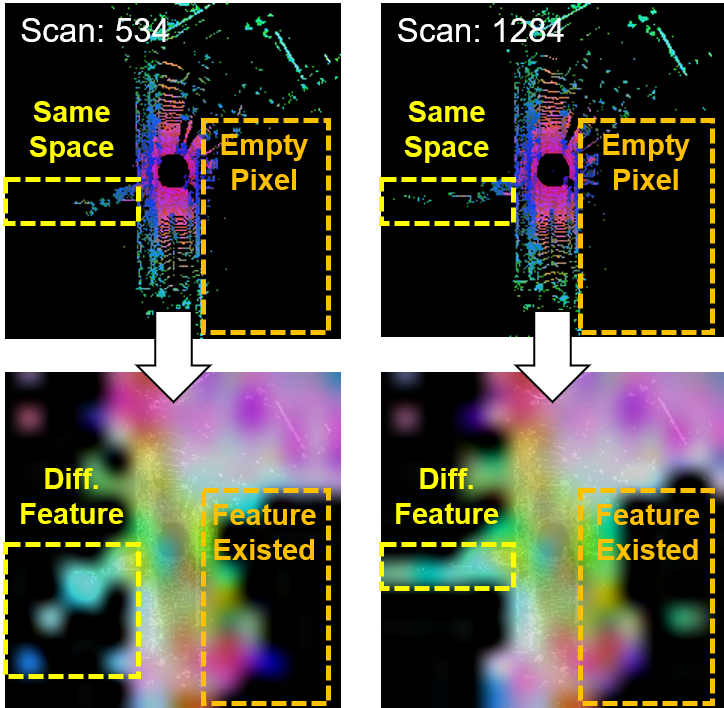}
        \label{fig:dino_bev}
    }
    \subfigure[Feature Visualization with RIV]{
        \includegraphics[width=.57\columnwidth]{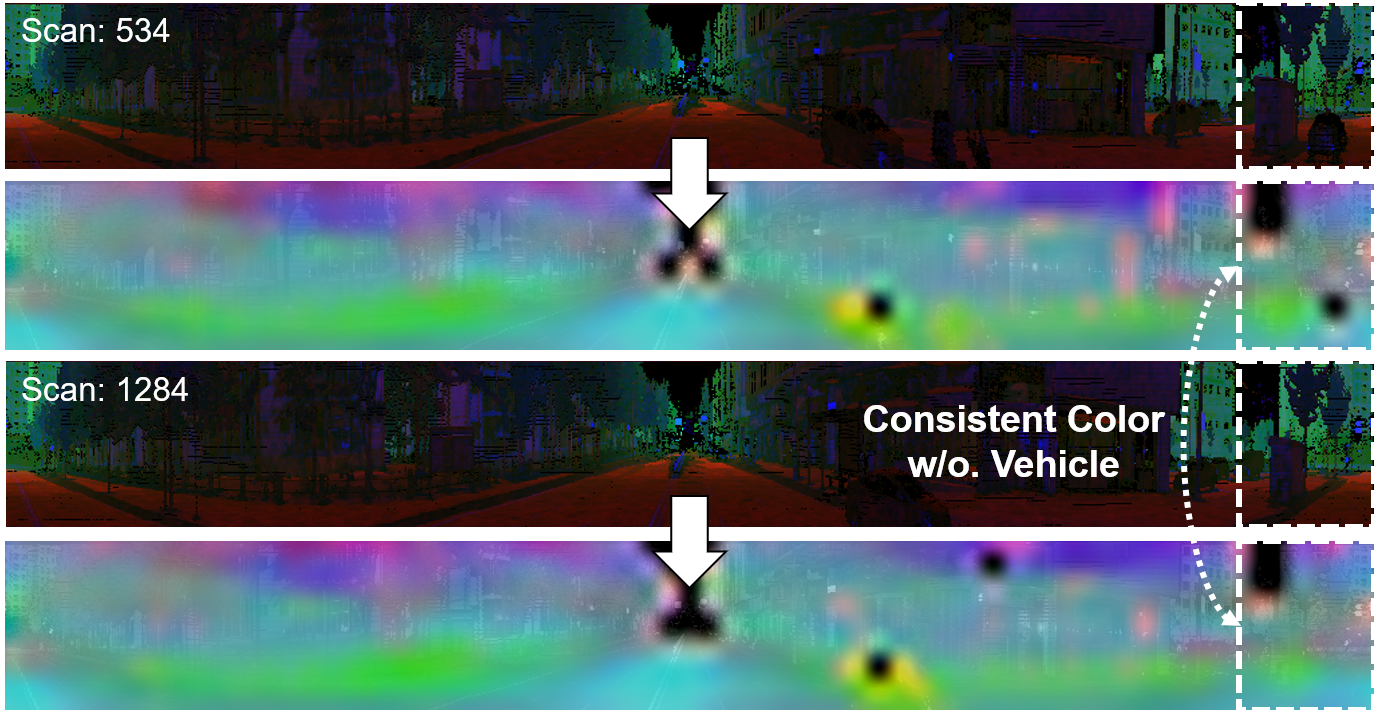}
        \label{fig:dino_riv}
    }
    \vspace{-3mm}
    \caption{Feature visualizations comparing BEV and RIV from \texttt{Roundabout01-O} positives. (a) BEV distorts feature shapes (yellow) and misidentifies empty pixels as features (orange), causing inconsistency. (b) RIV shows consistent features between scans, even with a missing car (white).}
    \label{fig:overall_dino}
    \vspace{-7mm}
\end{figure}

\textbf{BEV and RIV Representations with Patch-InfoNCE Loss: }We evaluate the effectiveness of BEV and RIV representations using three model variants: BEV without Patch-InfoNCE loss (\texttt{ExpB-1}), RIV without the loss (\texttt{ExpB-2}), and RIV with the loss (\texttt{ExpB-4}), since Patch-InfoNCE is not compatible with BEV. As shown in \tabref{tab:imlpr_comparison}, RIV consistently outperforms BEV, even without Patch-InfoNCE, and further improves with it by enhancing patch-level learning. Visualizations in \figref{fig:overall_dino} illustrate these differences. BEV struggles with feature representation: the yellow box shows distorted DINO features—diagonal in scan 534 and linear in scan 1284. The orange box highlights empty pixels misidentified as foreground, revealing DINOv2's limitations. In contrast, RIV produces consistent features across both scans. In \figref{fig:dino_riv}, white boxes remain stable even as a car disappears in scan 1284, with the region correctly excluded. These results indicate that \ac{VFM} performs more reliably with RIV, which better retains spatial and semantic patterns.

\textbf{MultiConv Adapter: }\tabref{tab:imlpr_comparison} shows the benefit of incorporating a MultiConv adapter (\texttt{ExpB-3}) versus not using one (\texttt{ExpB-4}). Even when the same number of last transformer blocks are fine-tuned, the presence of the adapter leads to significant performance improvements, notably increasing both average AR@1 and AF1. This confirms the adapter’s effectiveness in addressing sensor domain shifts while retaining the generalizable representations learned during pre-training. Additional and detailed ablation studies are provided in the appendix.

\input{tab/input_comparison}

%% file: tab/intra_results.tex
\begin{table}[t]
\centering
\caption{Performance Comparison in Intra-Session Place Recognition}
\label{tab:intra}
\resizebox{.95\textwidth}{!}{%
\begin{tabular}{l||cc|cc|cc|cc|cc|cc||cc}
\toprule
\multirow{2}{*}{Method} & 
\multicolumn{2}{c|}{\texttt{Round01-O}} & 
\multicolumn{2}{c|}{\texttt{Round02-O}} & 
\multicolumn{2}{c|}{\texttt{Round03-O}} & 
\multicolumn{2}{c|}{\texttt{Town01-O}} & 
\multicolumn{2}{c|}{\texttt{Town02-O}} & 
\multicolumn{2}{c||}{\texttt{Town03-O}} & 
\multicolumn{2}{c}{Average} \\ 
& R@1 & F1 
& R@1 & F1 
& R@1 & F1 
& R@1 & F1 
& R@1 & F1 
& R@1 & F1 
& R@1 & F1 \\ 
\midrule
LoGG3D-Net 
 & 0.425 & 0.740 
 & 0.257 & 0.475 
 & 0.542 & 0.806 
 & 0.101 & 0.274 
 & 0.135 & 0.256 
 & 0.151 & 0.299 
 & 0.269 & 0.475 \\
MinkLoc3d v2 
 & 0.933 & 0.968 
 & 0.856 & 0.937 
 & 0.908 & 0.959 
 & 0.847 & 0.925 
 & 0.834 & 0.922 
 & 0.907 & 0.957 
 & 0.881 & 0.945 \\
CASSPR 
 & 0.940 & 0.969 
 & 0.883 & 0.943 
 & 0.942 & 0.972 
 & 0.803 & 0.892 
 & 0.789 & 0.882 
 & 0.862 & 0.935 
 & 0.870 & 0.932 \\
BEVPlace++ 
 & \underline{0.975} & \underline{0.988} 
 & \underline{0.926} & \underline{0.962} 
 & \underline{0.962} & \underline{0.983} 
 & \underline{0.946} & \underline{0.975} 
 & \underline{0.942} & \underline{0.973} 
 & \underline{0.961} & \underline{0.980} 
 & \underline{0.952} & \underline{0.977} \\
ImLPR 
 & \textbf{0.992} & \textbf{0.996} 
 & \textbf{0.974} & \textbf{0.988} 
 & \textbf{0.992} & \textbf{0.997} 
 & \textbf{0.981} & \textbf{0.990} 
 & \textbf{0.950} & \textbf{0.977} 
 & \textbf{0.966} & \textbf{0.984} 
 & \textbf{0.976} & \textbf{0.989} \\
\bottomrule
\end{tabular}%
}
\small
\vspace{-0.1mm}
{$\,$}

\scriptsize 
\raggedright 
\, \, \, \, (\texttt{Round:} \texttt{Roundabout}, \textbf{Bold}: Best and \underline{Underline}: Second-Best
\vspace{-4mm}
\end{table}

%% file: tab/inter_roundabout.tex
\begin{table}[t!]
\centering
\caption{Performance Comparison in Inter-Session Place Recognition (HeLiPR \texttt{Roundabout-O})}
\label{tab:inter_roundabout}
\resizebox{.95\textwidth}{!}{%
\begin{tabular}{l||cc|cc|cc|cc|cc|cc||cc} 
\toprule
\multirow{2}{*}{Method} 
 & \multicolumn{2}{c|}{\texttt{Round01-02}} 
 & \multicolumn{2}{c|}{\texttt{Round01-03}} 
 & \multicolumn{2}{c|}{\texttt{Round02-01}} 
 & \multicolumn{2}{c|}{\texttt{Round02-03}} 
 & \multicolumn{2}{c|}{\texttt{Round03-01}} 
 & \multicolumn{2}{c||}{\texttt{Round03-02}} 
 & \multicolumn{2}{c}{Average} \\ 
 & R@1 & F1 & R@1 & F1 & R@1 & F1 & R@1 & F1 & R@1 & F1 & R@1 & F1 & R@1 & F1 \\ 
\midrule
LoGG3D-Net 
 & 0.610 & 0.847 
 & 0.610 & 0.847 
 & 0.577 & 0.801 
 & 0.637 & 0.844 
 & 0.617 & 0.855 
 & 0.681 & 0.859 
 & 0.622 & 0.842 \\
MinkLoc3Dv2 
 & 0.930 & \underline{0.970} 
 & 0.955 & \underline{0.981} 
 & 0.931 & 0.968 
 & 0.954 & 0.979 
 & 0.965 & \underline{0.985} 
 & 0.952 & 0.981 
 & 0.948 & 0.977 \\
CASSPR 
 & 0.939 & \underline{0.970} 
 & 0.961 & \underline{0.981} 
 & 0.945 & \underline{0.972} 
 & {0.968} & \underline{0.985} 
 & 0.962 & 0.982 
 & 0.964 & \underline{0.986} 
 & 0.957 & \underline{0.979} \\
BEVPlace++ 
 & \underline{0.963} & 0.918 
 & \underline{0.970} & 0.947 
 & \underline{0.956} & 0.948 
 & \underline{0.973} & 0.963 
 & \underline{0.979} & 0.964 
 & \underline{0.970} & 0.949 
 & \underline{0.969} & 0.948 \\
ImLPR 
 & \textbf{0.984} & \textbf{0.992} 
 & \textbf{0.991} & \textbf{0.996} 
 & \textbf{0.989} & \textbf{0.995} 
 & \textbf{0.991} & \textbf{0.996} 
 & \textbf{0.994} & \textbf{0.997} 
 & \textbf{0.992} & \textbf{0.997} 
 & \textbf{0.990} & \textbf{0.996} \\ 
\bottomrule
\end{tabular}%
}

\vspace{-5mm}
\end{table}

%% file: tab/inter_town.tex
\begin{table}[t!]
\centering
\caption{Performance Comparison in Inter-Session Place Recognition (HeLiPR \texttt{Town-O})}
\label{tab:inter_town}
\resizebox{.95\textwidth}{!}{%
\begin{tabular}{l||cc|cc|cc|cc|cc|cc||cc}
\toprule
\multirow{2}{*}{Method}
 & \multicolumn{2}{c|}{\texttt{Town01-02}}
 & \multicolumn{2}{c|}{\texttt{Town01-03}}
 & \multicolumn{2}{c|}{\texttt{Town02-01}}
 & \multicolumn{2}{c|}{\texttt{Town02-03}}
 & \multicolumn{2}{c|}{\texttt{Town03-01}}
 & \multicolumn{2}{c||}{\texttt{Town03-02}}
 & \multicolumn{2}{c}{Average} \\
 & R@1 & F1 & R@1 & F1 & R@1 & F1 & R@1 & F1 & R@1 & F1 & R@1 & F1 & R@1 & F1 \\
\midrule
LoGG3D-Net &
  0.282 & 0.546 &
  0.281 & 0.478 &
  0.280 & 0.538 &
  0.332 & 0.568 &
  0.272 & 0.482 &
  0.330 & 0.587 &
  0.296 & 0.533 \\
MinkLoc3Dv2 &
  {0.948} & \underline{0.975} &
  {0.962} & \underline{0.981} &
  {0.960} & \underline{0.980} &
  {0.958} & \underline{0.980} &
  {0.950} & \underline{0.976} &
  {0.960} & 0.980 &
  {0.956} & \underline{0.979} \\
CASSPR &
  0.916 & 0.958 &
  0.912 & 0.957 &
  0.939 & 0.970 &
  0.947 & 0.975 &
  0.924 & 0.963 &
  0.934 & 0.967 &
  0.929 & 0.965 \\
BEVPlace++ &
  \underline{0.960} & 0.954 &
  \underline{0.966} & 0.960 &
  \underline{0.974} & 0.953 &
  \underline{0.983} & 0.972 &
  \underline{0.971} & 0.972 &
  \underline{0.982} & \underline{0.989} &
  \underline{0.973} & 0.967 \\
ImLPR &
  \textbf{0.989} & \textbf{0.995} &
  \textbf{0.991} & \textbf{0.996} &
  \textbf{0.988} & \textbf{0.996} &
  \textbf{0.988} & \textbf{0.994} &
  \textbf{0.985} & \textbf{0.993} &
  \textbf{0.990} & \textbf{0.995} &
  \textbf{0.989} & \textbf{0.995} \\
\bottomrule

\end{tabular}%
}
\vspace{-3.5mm}
\end{table}

%% file: tab/input_comparison.tex
\begin{table}[t]
\centering
\caption{Ablation Study on Input Representation, Loss Function, and Adapter for Inter-Session \ac{LPR}}
\label{tab:imlpr_comparison}
\resizebox{.95\columnwidth}{!}{%
\begin{tabular}{l||ccc||cc|cc|cc|cc|cc}
\toprule
\multirow{2}{*}{Method} & 
  \multirow{2}{*}{Input} &
  \multirow{2}{*}{Adapter} &
  \multirow{2}{*}{Loss} &
  \multicolumn{2}{c|}{\texttt{Roundabout-O}} & 
  \multicolumn{2}{c|}{\texttt{Town-O}} & 
  \multicolumn{2}{c|}{\texttt{DCC}} & 
  \multicolumn{2}{c|}{\texttt{Roundabout-V}} & 
  \multicolumn{2}{c}{Average} \\ 
 & & & & AR@1 & AF1 & AR@1 & AF1 & AR@1 & AF1 & AR@1 & AF1 & AR@1 & AF1 \\ 
\midrule
\texttt{ExpB-1} & BEV & \cmark & $\mathcal{L}_{\text{TSAP}}$ & 0.945 & 0.977 & 0.923 & 0.968 & 0.901 & 0.966 & \underline{0.869} & \underline{0.942} & 0.911 & 0.965 \\
\texttt{ExpB-2} & RIV & \cmark& $\mathcal{L}_{\text{TSAP}}$ & \underline{0.979} & \underline{0.990} & \underline{0.985} & \underline{0.993} & \textbf{0.945} & \underline{0.970} & 0.844 & 0.920 & \underline{0.938} & \underline{0.968} \\
\texttt{ExpB-3} & RIV & \xmark & $\mathcal{L}_{P}, \mathcal{L}_{\text{TSAP}}$ & 0.850 & 0.925 & 0.833 & 0.913 & 0.850 & 0.943 & 0.645 & 0.809 & 0.795 & 0.898 \\
\texttt{ExpB-4} & RIV & \cmark & $\mathcal{L}_{P}, \mathcal{L}_{\text{TSAP}}$ & \textbf{0.990} & \textbf{0.996} & \textbf{0.989} & \textbf{0.995} & \underline{0.942} & \textbf{0.973} & \textbf{0.888} & \textbf{0.948} & \textbf{0.952} & \textbf{0.978} \\
\bottomrule
\end{tabular}%
}
\vspace{-5mm}
\end{table}

%% file: contents/5_conclusion.tex
 \section{Conclusion}
\label{sec:conclusion}
In this paper, we introduced ImLPR, the first novel pipeline that adapts the DINOv2 \ac{VFM} to bridge the LiDAR and vision domains. The proposed approach utilized a RIV representation with three channels, integrated MultiConv adapters, and employed Patch-InfoNCE loss for patch-level contrastive learning. Evaluations on the public datasets demonstrate that ImLPR outperforms SOTA methods in both intra-session and inter-session \ac{LPR}, achieving superior performance. Our approach redefines \ac{LPR} by moving beyond traditional 3D sparse convolution and BEV-based descriptors. Instead, we leverage \ac{VFM} with RIV images, rather than BEV, for enhanced feature representation.
Future work could explore integrating ImLPR with other foundation models, such as segmentation models, or developing hierarchical pipelines. These approaches could combine ImLPR’s patch-based retrieval with a segmentation-driven re-ranking process, where local patch correspondences are refined to improve matching accuracy, enhancing overall \ac{LPR} performance.

%% file: contents/6_limitation.tex
\section{Limitation}
\label{sec:limitation}
While ImLPR achieves \ac{SOTA} results on a variety of datasets, its application remains limited to homogeneous \ac{LPR}, where the same type of LiDAR sensor is used for both query and database. It is not yet suited for the emerging task of heterogeneous \ac{LPR} \cite{jung2025helios}, which requires matching across different LiDAR sensor types with varying specifications and fields of view. Accumulating a batch of scans, as adopted in our experiments, partially mitigates this limitation by improving robustness across sessions, but significant discrepancies in sensor geometry prevent it from serving as a complete solution. Furthermore, although we fine-tuned the DINOv2 vision foundation model with MultiConv adapters to better align with the LiDAR domain, effectively addressing heterogeneous \ac{LPR} or broader generalization still demands training on larger and more diverse datasets. Despite these constraints, ImLPR’s \ac{VFM}-based approach extends beyond traditional place recognition methods by adapting pre-trained DINOv2 features to LiDAR RIV inputs. This \ac{VFM}-based approach enables future development of generalizable place recognition systems through integration with other foundation models, such as those for segmentation \cite{kirillov2023segment} and multi-modal processing \cite{radford2021learning}.

%% file: supplementary/supplementary.tex
\section{Experimental Setup}
\textbf{Implementation Details: }ImLPR is trained by fine-tuning the final two transformer blocks, with MultiConv adapters integrated every three blocks. LiDAR point clouds are converted into images with dimensions \(H \times W = 126 \times 1022\). The image channels comprise reflectivity, range, and normal ratio. Reflectivity acts as a semantic cue for scene segmentation. For LiDAR sensors without reflectivity data, the intensity channel is utilized during inference, serving a comparable semantic purpose. The singular value ratio is calculated using \(k=8\) nearest neighbors for the HeLiPR-O dataset and \(k=25\) for the MulRan, NCLT and HeLiPR-V datasets. To ensure continuity in cylindrical images, the leftmost and rightmost 28 columns (equivalent to 2 patches) are appended to the opposite sides, maintaining seamless connectivity during both training and inference. Feature aggregation uses parameters \((m, l, e) = (128, 64, 256)\) to construct the descriptor. For the Patch-InfoNCE loss, 192 positive and 128 negative patch pairs are sampled per image, with a temperature \(\tau_l = 0.2\). Negative patches are selected based on patch distance thresholds \(v_{\text{dist}} = 3\) and \(h_{\text{dist}} = 20\). The Patch-InfoNCE loss is computed using only 1/8 of the positive image pairs within the batch to optimize computational efficiency. To establish correspondence between two LiDAR scans represented as RIV images, the RIV images are first converted into 3D point clouds and voxelized at a 0.4-meter resolution, followed by precise alignment using the ICP alignment. For the TSAP loss, a temperature \(\tau_g = 0.01\) is applied, truncated to the top four ranked descriptors, with a batch size of 2048. The combined loss is weighted with \(\lambda = 2.0\). Training proceeds for 100 epochs using the AdamW optimizer, with a learning rate of \(5 \times 10^{-4}\), a 1/10 warmup phase, and a cosine scheduler. When performed on three NVIDIA GeForce RTX 3090 GPUs, this is completed in 12 hours.

\textbf{Evaluation Metrics:}  
We evaluate the performance using Recall@1 (R@1) and the maximum F1 score (F1). Recall@1 measures the percentage of queries where the top-1 retrieved scan is a correct match (within 10 meters of the ground truth), defined as:

\begin{equation}
    \text{Recall@1} = \frac{\text{Number of queries with correct top-1 match}}{\text{Total number of query candidates}}
\end{equation}

The F1 score represents the optimal harmonic mean of precision and recall across all thresholds, calculated as:
\begin{equation}
\text{F1 score} = 2 \times \frac{\text{Precision} \times \text{Recall}}{\text{Precision} + \text{Recall}}, \quad \text{Precision} = \frac{\text{TP}}{\text{TP} + \text{FP}}, \quad \text{Recall} = \frac{\text{TP}}{\text{TP} + \text{FN}}
\end{equation}
where \(\text{TP}\), \(\text{FP}\), and \(\text{FN}\) denote the number of True Positives, False Positives, and False Negatives for each threshold. The maximum F1 score is the highest F1 value obtained by varying the threshold for positive matches.  Additionally, we report average results, including Average Recall@1 (AR@1) and Average maximum F1 score (AF1), computed as the mean values across multiple sequences to provide a comprehensive assessment of ImLPR’s performance.

\section{Truncated SmoothAP Loss with Global Image Descriptor}
To train ImLPR with the global descriptor, we adopt the Truncated SmoothAP (TSAP) loss \cite{komorowski2022improving}. Unlike the triplet and contrastive loss, TSAP optimizes Average Precision (AP) through a continuous approximation, focusing ranking evaluation on top candidates to reduce computation. For a query descriptor \(g_q\) with positives in \(\mathcal{P}^+\) and all batch descriptors in \(\mathcal{D}\), TSAP is formulated as:
\begin{align}
\small
\mathcal{L}_{\text{TSAP}} = \frac{1}{m}\sum_{q=1}^{m}\bigl(1-\text{AP}_q\bigr), \:\:\: \text{AP}_q = \frac{1}{|\mathcal{P}^+|} \sum_{i \in \mathcal{P}^+} \frac{1 + \sum_{j \in \mathcal{P}^+, j \neq i} H\bigl(d_g(q, i) - d_g(q,j); \tau_g\bigr)}
{1 + \sum_{j \in \mathcal{D}, j \neq i} H\bigl(d_g(q, i) - d_g(q,j); \tau_g\bigr)},
\normalsize
\end{align}
where \(H\bigl(x; \tau_g\bigr) = \bigl(1 + e^{x/\tau_g}\bigr)^{-1}\) approximates a step function, with \(\tau_g\) as a temperature parameter. Summation over \(\mathcal{D}\) uses a top-ranked subset to reduce computation. In large-batch training, TSAP matches retrieval metrics such as Average Recall, boosting global descriptor performance. We also apply multi-staged backpropagation \cite{revaud2019learning} for efficient large-batch optimization.

\section{Data Augmentation in ImLPR}

To enhance the robustness of ImLPR, \ac{RIV} images are subjected to various data augmentations during training. Images are randomly rotated through yaw, sampled uniformly from 0 to \(2\pi\) radians, using cyclic horizontal column shifts. This promotes robustness to orientation variations prevalent in LiDAR scans. The reflectivity and normal ratio channels are scaled to the [0, 1] range by dividing by 255, while the range channel is normalized by dividing by the LiDAR's maximum range, ensuring uniform input scales.

\begin{figure}[!t]
    \centering
    {
    \includegraphics[trim = 0cm 0.5cm 0cm 0.5cm, width=.9\columnwidth]{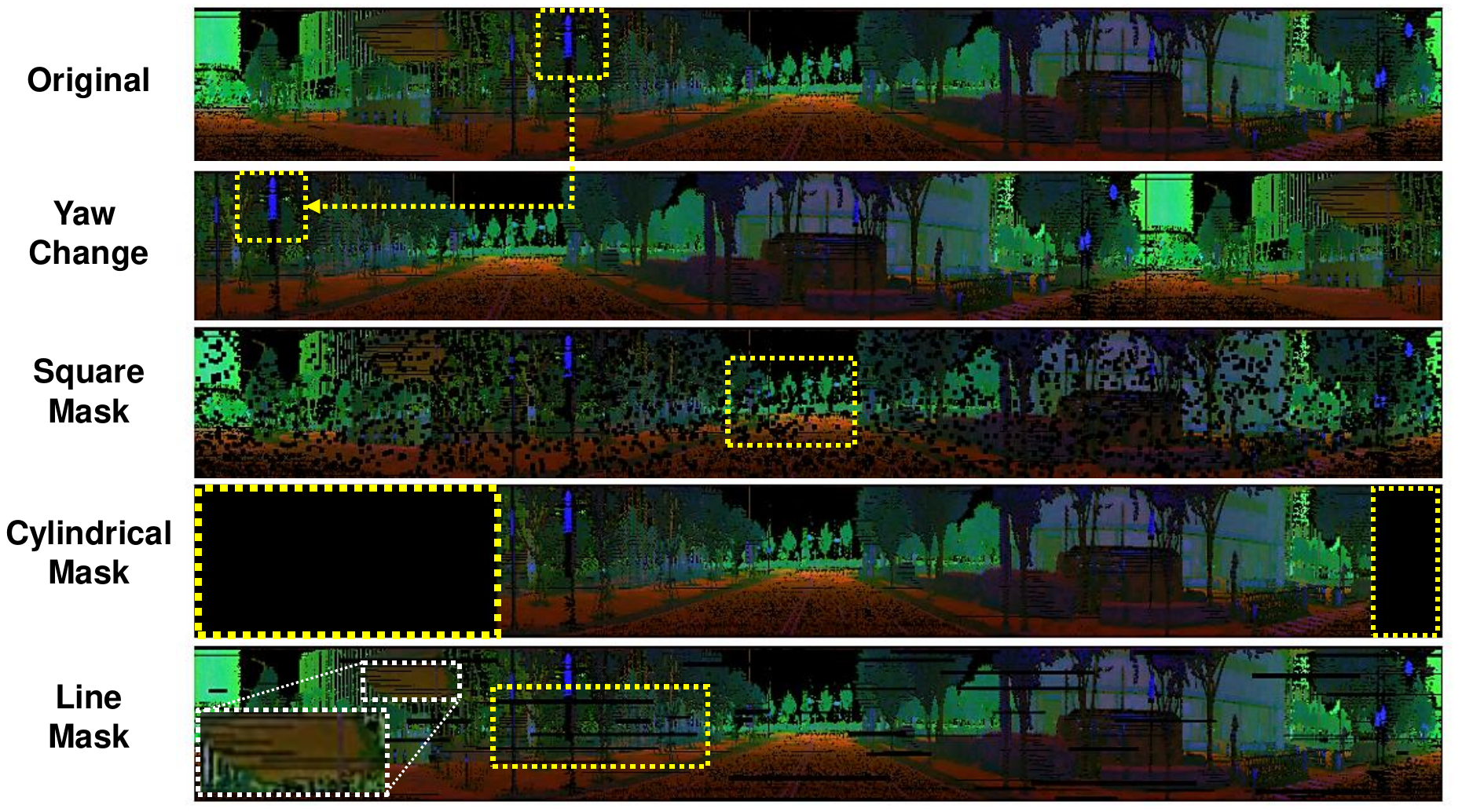}
        
    }
    \caption{Augmented image featuring yaw variation, square mask, cylindrical mask, and line mask. The yellow line represents a leftward sign displacement, corresponding to horizontal pixel shifts induced by yaw variation. Yellow boxes emphasize augmented areas, highlighting deviations from the original image. In the line mask, the white box indicates noise from vacant line pixels caused by point cloud projection, whereas our yellow box showcases augmentation similar to the original.}
    \label{fig:augmentation}
    \vspace{-6mm}
\end{figure}

Three masking strategies are employed, as illustrated in \figref{fig:augmentation}. Random patch masking applies square patches of varying sizes, occluding up to 40\% of the image area based on a randomly determined mask ratio. Cylindrical masking introduces a contiguous mask, spanning up to 30\% of the image width, with random starting positions and cylindrical wrapping to account for boundary effects. These techniques address challenges such as occlusions and sparse scene representations. Line-style masking incorporates randomly placed rectangular lines to emulate projection artifacts, where multiple points collapse into a single pixel or horizontal lines appear vacant due to RIV projection. These augmentations bolster ImLPR's robustness, significantly contributing to its superior performance in both intra-session and inter-session place recognition.

\section{Datasets}
\label{sec:dataset}
To evaluate ImLPR, we utilize three public datasets: HeLiPR, NCLT, and MulRan. An example scan from each dataset, represented as a \ac{RIV}, is depicted in \figref{fig:dataset_img}.

\subsection{HeLiPR Dataset}
The HeLiPR dataset comprises six distinct environments---\texttt{Roundabout01-03}, \texttt{Town01-03}, \texttt{Bridge01-04}, \texttt{DCC04-06}, \texttt{KAIST04-06}, and \texttt{Riverside04-06}---captured using four LiDAR sensors: Ouster OS2-128, Velodyne VLP-16C, Livox Avia, and Aeva Aeries II. Each sequence covers approximately 8.5 km, providing sufficient coverage for identifying multiple place recognition pairs. For training, we employ the Ouster OS2-128 sensor across the \texttt{DCC04-06}, \texttt{KAIST04-06}, and \texttt{Riverside04-06} sequences, yielding a total of 16,435 scans. The test set includes the \texttt{Roundabout01-03} sequences from both the Ouster OS2-128 and Velodyne VLP-16C sensors, and the \texttt{Town01-03} sequences from only the Ouster OS2-128 sensor, with a total of 15,375 scans for the Ouster OS2-128 and 7,728 scans for the Velodyne VLP-16C. Sequences are labeled as \texttt{Sequence-Sensor} (e.g., \texttt{Roundabout01-O} for Ouster, \texttt{Roundabout01-V} for Velodyne) to distinguish sensor-specific data.

\begin{figure}[!t]
    \centering
    \subfigure[Input image from each dataset]{
    \includegraphics[width=.45\columnwidth]{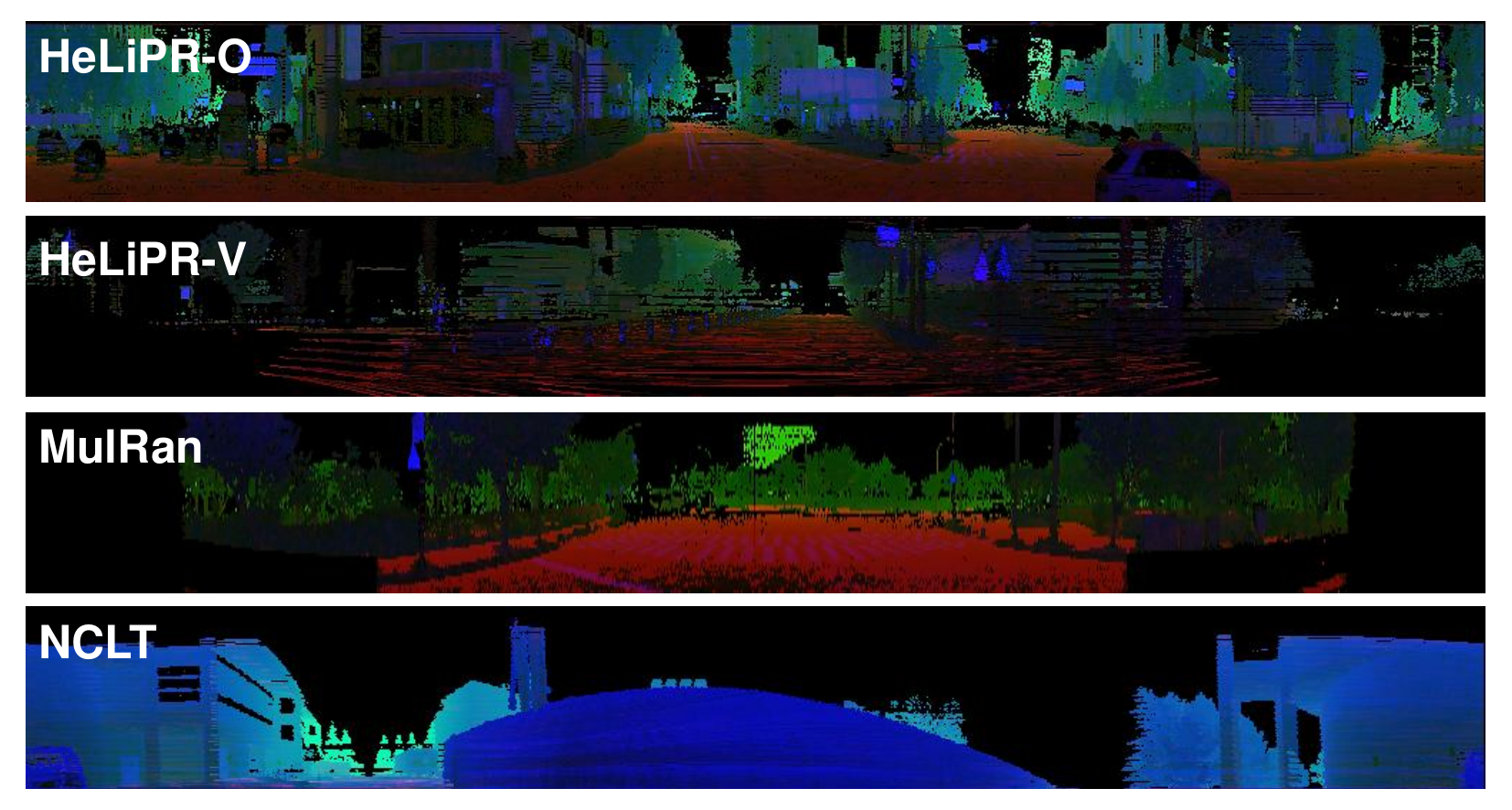}
        \label{fig:dataset_img}
    }
    \subfigure[Intensity distribution per scan]{
        \includegraphics[trim = 0cm 0.5cm 0cm 0.5cm, width=.5\columnwidth]{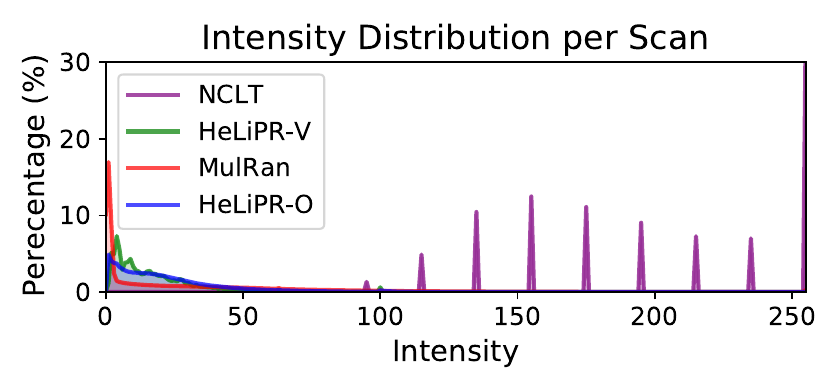}
        \label{fig:intensity_dist}
    }
    \vspace{-2mm}
    \caption{(a) Visualization of RIV images from four datasets across various LiDARs. HeLiPR with Ouster (HeLiPR-O) delivers high-resolution point clouds with minimal empty pixels. MulRan exhibits pronounced empty regions at the leftmost and rightmost edges, attributed to sensor occlusion. Similarly, HeLiPR with Velodyne (HeLiPR-V) suffers from sparse point clouds, resulting in persistent empty pixels despite scan accumulation. In contrast, NCLT scans, benefiting from slower platform motion, display few empty pixels; however, their intensity distribution is markedly distinct, characterized by discrete and inverted values, yielding a predominantly blue appearance. (b) Intensity distributions for each scans from HeLiPR, MulRan, and NCLT, with NCLT demonstrating a significant distributional difference.}
    \label{fig:dataset}
    \vspace{-6mm}
\end{figure}

For the \texttt{Roundabout-V} sequence, we aggregate 5 seconds of scan data to form a submap, which is subsequently projected into a RIV image. To avoid redundant accumulation of static scans, we exclude scans acquired during stationary periods, retaining only those captured during motion. This strategy ensures consistent vertical image dimensions, partially mitigating the challenges posed by sparse point clouds in lower-dimensional representations. However, as shown in \figref{fig:dataset_img}, the inherent sparsity of the LiDAR data results in persistent empty pixels, making this dataset a challenging testbed for evaluating the robustness of \ac{LPR} methods to incomplete representations.

\subsection{MulRan Dataset}

The MulRan dataset covers four urban environments---\texttt{DCC}, \texttt{KAIST}, \texttt{Riverside}, and \texttt{Sejong}---each including three sequences (\texttt{01-03}) acquired with an Ouster OS1-64 LiDAR. For this dataset, we focus exclusively on the \texttt{DCC01-03} sequences, which span an average distance of 4.9km. The dataset poses challenges due to the lower vertical resolution of this LiDAR sensor and occlusions induced by a secondary sensor positioned behind the LiDAR, leading to $20\%$ of RIV image pixels remaining empty, as shown in \figref{fig:dataset_img}. Point clouds are projected into RIV images of dimension $1022 \times 64$ and subsequently resized to $1022 \times 128$ via linear interpolation. Given that MulRan intensity values exceed 255, we apply normalization to ensure compatibility. A total of 4,328 scans are utilized as test sets for generalization experiments and ablation studies.

\subsection{NCLT Dataset}
The NCLT dataset comprises 27 sequences recorded in a college campus using a Velodyne HDL-32E LiDAR mounted on a Segway platform. Owing to the sparse nature of the point clouds, we aggregate scans using the methodology applied to HeLiPR-V dataset. The Segway’s reduced velocity, relative to the vehicle used for HeLiPR, yields RIV images with fewer empty pixels, closely resembling the high-resolution images of the HeLiPR-O dataset. Nevertheless, the intensity distribution markedly differs from other datasets, exhibiting discrete values and a prevalence of high-intensity points, in contrast to the predominantly low-intensity points typical of other datasets. This difference results in blurrier NCLT RIV images as depicted in \figref{fig:dataset}. To mitigate this, we invert and rescale the intensity distribution. Despite these adjustments, the discrete and inaccurate intensity channel continues to impair the semantic quality of RIV images, posing challenges for \ac{LPR}. For generalization testing, we utilize the first three sequences---\texttt{2012-01-08}, \texttt{2012-01-15}, and \texttt{2012-01-22}---encompassing a total of 6,368 scans as test sets.

\begin{figure}[!t]
    \centering
    {
    \includegraphics[trim = 0cm 0cm 0cm 0cm, width=\columnwidth]{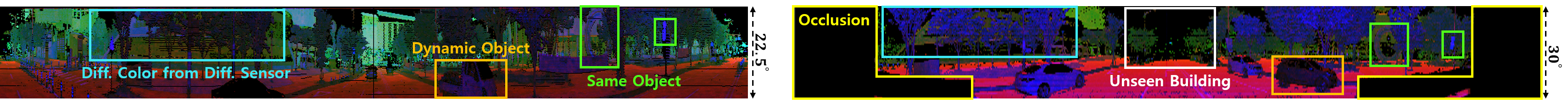}
        
    }
    \caption{RIV images from HeLiPR (Left) and MulRan (Right) DCC sequence, taken 0.3m apart, vary in appearance: occluded area (sensor location), hidden buildings (range), warped proportions (FoV and sensor location), and shifted colors (intensity distributions).}
    \label{fig:place_difference}
    
\end{figure}

\begin{figure}[!t]
    \vspace{-9.5mm}
    \centering
    \begin{minipage}{0.58\textwidth}  
        \centering
        \includegraphics[trim = 0cm 0cm 0cm 0cm, width=\linewidth]{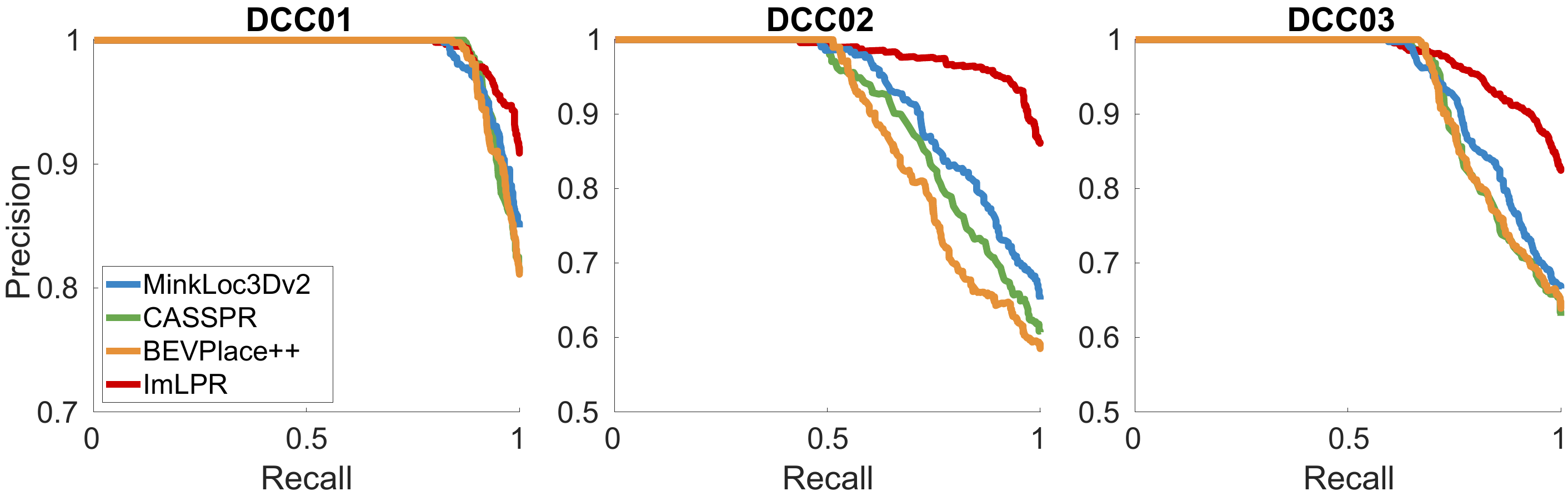}
        \vspace{-9.5mm}
    \end{minipage}%
    \hfill
    \begin{minipage}{0.41\textwidth}  
        \centering   
        \begin{table}[H]
            \caption{Results in MulRan dataset}
            \centering
            \resizebox{\linewidth}{!}{
                \begin{tabular}{l||cc|cc|cc|cc}
\toprule
 & \multicolumn{8}{c}{MulRan \texttt{DCC} (OS1-64)} \\ 
\cline{2-9}
\multirow{1}{*}{Method} & 
\multicolumn{2}{c|}{01} & 
\multicolumn{2}{c|}{02} & 
\multicolumn{2}{c|}{03}  & 
\multicolumn{2}{c}{Average}\\ 
 & R@1 & F1 & R@1 & F1 & R@1 & F1 & AR@1 & AF1 \\ 
\midrule
LoGG3D-Net (D) & 0.117 & 0.310 & 0.075 & 0.144 & 0.090 & 0.187 & 0.094 & 0.214\\
CASSPR (D) & 0.813 & {0.943} & 0.607 & 0.793 & 0.629 & 0.818 & 0.683& 0.851\\ \midrule
MinkLoc3dv2 (D)& {0.849} & 0.939 & {0.651} & {0.829} & {0.666} & {0.841} & {0.722} & {0.870}\\
MinkLoc3Dv2 (R)& {0.853}  & {0.932} & {0.632} & {0.814} & {0.688} & {0.832} & {0.724} & {0.859} \\ \midrule
BEVPlace++ (D)& {0.811}  & {0.937} & {0.585} & {0.766} & {0.639} & {0.813} & {0.678} & {0.839} \\ 
BEVPlace++ (R)& {0.822}  & {0.922} & {0.588} & {0.773} & {0.655} & {0.811} & {0.688} & {0.835} \\ \midrule 
ImLPR (D)& \underline{0.909}  & \underline{0.965} & \textbf{0.861} & \textbf{0.945} & \textbf{0.825} & \underline{0.918}  & \textbf{0.865} & \underline{0.943} \\
ImLPR (R)& \textbf{0.918}  & \textbf{0.968} & \underline{0.844} & \underline{0.940} & \textbf{0.825} & \textbf{0.925} & \underline{0.862} & \textbf{0.944}\\
\bottomrule
\end{tabular}%

            }
                \vspace{6mm}           
            \label{tab:generalized_dcc}
        \end{table}
    \end{minipage}
    \vspace{-4mm}
    \caption{Performance of High-Resolution LiDAR-Trained Models on MulRan  dataset (OS1-64)}
    \vspace{-5mm}
    \label{fig:generalized_dcc}
\end{figure}

\section{Evaluation on Generalization Capability}
Following the summarized generalization experiments in the main paper, here we further assess the performance of the trained model in detail using three datasets. Consistent with the primary evaluation, we conduct intra-session place recognition using the HeLiPR-O trained model without additional fine-tuning. Scan accumulation follows the methodology outlined in \secref{sec:dataset}. To ensure a fair comparison, all methods are assessed using the same accumulated scans.

\subsection{Evaluation on MulRan Dataset}
We evaluate the generalization ability of ImLPR on the MulRan \texttt{DCC01-03} sequences. Although the training and test sets have spatial overlap, they differ substantially due to a four-year temporal gap, differences in LiDAR hardware (OS1-64), and environmental changes such as occlusions. These differences are illustrated in \figref{fig:place_difference}. For evaluation, LiDAR scans are projected into $1022\times64$ RIV images and resized to $1022\times126$ to normalize vertical ray differences.

To ensure the robustness of our evaluation protocol and rule out the possibility of performance inflation due to overlap, we additionally replace the original training sequences (\texttt{DCC04-06}) with a geographically seperate sequence, \texttt{Roundabout04-06}, from the HeLiPR with Ouster. We retrain three representative methods, MinkLoc3Dv2 (point-based), BEVPlace++ (BEV-based), and ImLPR (RIV-based), on both training sets: the original (denoted as D) and the revised one (denoted as R).

As shown in \tabref{tab:generalized_dcc} and \figref{fig:generalized_dcc}, all methods exhibit only minor differences in performance between the two training sets, validating that the evaluation results are not overly dependent on spatial overlap. Moreover, all models experience performance degradation on the MulRan dataset compared to high-resolution training data, highlighting the impact of domain shift. Among them, ImLPR consistently achieves the highest performance, regardless of training configuration.
These results demonstrate that ImLPR generalizes robustly across datasets with different LiDAR sensors and environmental conditions. MinkLoc3Dv2 benefits from sparse convolution that avoids feature generation in occluded regions, ranking second. BEVPlace++ ranks lower, as BEV images tend to retain features even in empty or occluded areas. ImLPR, by contrast, effectively handles low resolution and occlusions through its robust RIV representation and powerful feature extractor. These results demonstrate that ImLPR exhibits strong generalization on the MulRan dataset, even under substantial domain shifts.

\subsection{Evaluation on NCLT Dataset}

\begin{figure}[!t]
    \centering
    {
    \includegraphics[trim = 0cm 0cm 0cm 0cm, width=\columnwidth]{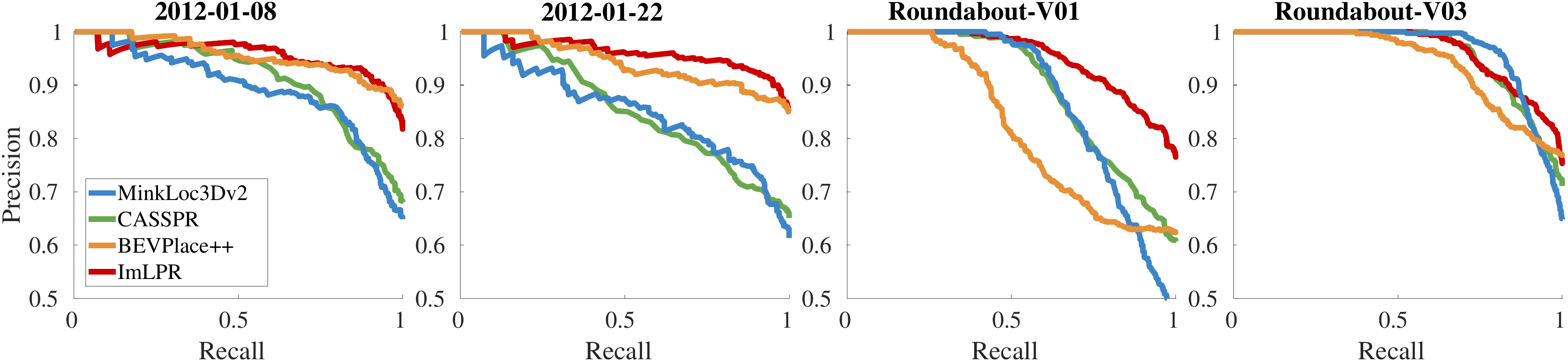}
    }
    \vspace{-5mm}
    \caption{Precision-Recall curves for generalization assessment on NCLT and HeLiPR-V datasets. ImLPR demonstrates robust and consistent performance across both datasets. Conversely, BEVPlace++ yields strong results on NCLT but diminished performance on HeLiPR \texttt{Roundabout-V}. Sparse 3D convolution methods, MinkLoc3Dv2 and CASSPR, exhibit variable performance across the evaluated datasets.}
    \label{fig:generalization}
    \vspace{-5mm}
\end{figure}

As reported in \tabref{tab:sensor_comparison}, ImLPR and BEVPlace++ exhibit comparable performance across the evaluated datasets. This similarity stems from the discrete and inaccurate intensity values, which diminish pixel-level distinctions in RIV images and hinder semantic interpretation. Nevertheless, ImLPR surpasses 3D sparse convolution methods, such as MinkLoc3Dv2, by leveraging geometric information from the range and normal ratio channels effectively. Although ImLPR’s Recall@1 and F1 score for the \texttt{2012-01-08} sequence are marginally lower than those of BEVPlace++, the Precision-Recall curve, shown in \figref{fig:generalization}, reveals that ImLPR remains competitive with BEVPlace++ and achieves a superior \ac{AUC} for the \texttt{2012-01-22} sequence. Furthermore, despite the unreliable intensity channel, ImLPR attains the highest average F1 score and the second-highest Average Recall@1, closely following BEVPlace++. These findings highlight ImLPR’s robust generalization capabilities and its efficacy in facilitating \ac{LPR} under challenging intensity conditions.

\input{tab/gen_dcc}

\subsection{Evaluation on HeLiPR-V Dataset}

ImLPR surpasses competing methods across nearly all of our evaluation metrics, securing the highest Recall@1 and F1 score for the \texttt{Roundabout01-V} and \texttt{Roundabout02-V} sequences, and delivering competitive performance in \texttt{Roundabout03-V}, as detailed in \tabref{tab:sensor_comparison}. On average, ImLPR achieves a performance improvement more than $10\%$ better than the second-ranked method, BEVPlace++, in both Recall@1 and F1 score. In contrast, other methods show significant performance variability. This is primarily due to the different point cloud distributions which were unobserved during training, which greatly hinder their generalization. Notably, the second-highest F1 scores across the three HeLiPR-V sequences are inconsistent, each attained by a different method, highlighting the difficulty that existing methods have in achieving reliable performance. The Precision-Recall curves presented in \figref{fig:generalization} indicate that for \texttt{Roundabout03-V}, ImLPR and MinkLoc3Dv2 yield comparable \ac{AUC}, followed by CASSPR and BEVPlace++. For \texttt{Roundabout01-V}, however, ImLPR markedly outperforms other methods, with CASSPR, MinkLoc3Dv2, and BEVPlace++ trailing in that order. These results underscore ImLPR’s consistently high performance across the HeLiPR-V dataset.

Despite performance degradation across the MulRan, NCLT, and HeLiPR-V datasets due to domain shifts, sensor disparities, inaccurate intensity values, and occlusions, ImLPR demonstrates robust performance. This resilience stems from two key factors. First, ImLPR’s adept fusion of geometric and semantic features included in RIV channels enables reliable \ac{LPR} even with low-resolution or unstable inputs. Second, its exceptional generalization performance, enabled by VFM, overcomes the limitations of traditional domain-specific training, which struggles to achieve generalizability. These observations underscore ImLPR’s potential as a robust framework for generalized \ac{LPR} across diverse and challenging environments.

\begin{table}[!t]
\centering
\caption{Intra-session PR with different threshold in \texttt{Roundabout01-O}}
\label{tab:radius}
\resizebox{\columnwidth}{!}{%
\begin{tabular}{l|cc|cc|cc|cc|cc} \toprule
 & \multicolumn{2}{c|}{Threshold 1m}& \multicolumn{2}{c|}{Threshold 3m}                & \multicolumn{2}{c|}{Threshold 5m}                & \multicolumn{2}{c|}{Threshold 7m} & \multicolumn{2}{c}{Threshold 10m}               \\ 
Method & \multicolumn{1}{c}{R@1} & \multicolumn{1}{c|}{F1} & \multicolumn{1}{c}{R@1} & \multicolumn{1}{c|}{F1} & \multicolumn{1}{c}{R@1} & \multicolumn{1}{c|}{F1} & \multicolumn{1}{c}{R@1} & \multicolumn{1}{c|}{F1} & \multicolumn{1}{c}{R@1} & \multicolumn{1}{c}{F1} \\ \midrule
MinkLoc3Dv2 & 0.732 & 0.963 & 0.933 & 0.968 & 0.798 & 0.888 & 0.839 & 0.915 & 0.933 & 0.968\\
BEVPlace++   & \underline{0.913} & \underline{0.959} & \underline{0.904} & \underline{0.951} & \underline{0.909} & \underline{0.953} & \underline{0.921} & \underline{0.959} & \underline{0.975}  & \underline{0.988} \\
ImLPR & \textbf{0.965}  & \textbf{0.987} & \textbf{0.933} & \textbf{0.970} & \textbf{0.975} & \textbf{0.988} & \textbf{0.960} & \textbf{0.980} & \textbf{0.992}  &  \textbf{0.996}    \\ \bottomrule
\end{tabular}%
}
\vspace{-2mm}
\end{table}

\section{Performance Variations According to Different Thresholds}

To rigorously evaluate spatial precision in place recognition, we analyze the impact of varying the distance threshold used to define positive correspondences. While a default threshold of \unit{10}{\meter}, which roughly corresponds to the width of a four-lane road, is a reasonable setting, this choice often results in uniformly high scores across methods. limiting discriminative insight. We therefore systematically tighten the thresholds, and evaluate intra-session performance on \texttt{Roundabout01-Ouster} using three representative methods: MinkLoc3Dv2 (3D point cloud-based), BEVPlace++ (BEV image-based), and ImLPR (RIV image-based).

As presented in \tabref{tab:radius}, ImLPR exhibits consistently superior performance across all threshold levels. Notably, under the most stringent \unit{1}{\meter} setting, it achieves an R@1 of 0.965 and F1 score of 0.987, substantially outperforming both BEVPlace++ and MinkLoc3Dv2. This indicates that ImLPR is capable of not only recognizing the correct place but also localizing it with high spatial accuracy. BEVPlace++ maintains relatively stable performance across thresholds but shows limited gains under stricter criteria, highlighting its reduced sensitivity to fine-grained spatial alignment. In contrast, MinkLoc3Dv2 exhibits fluctuations due to the sparsity of true positives at lower thresholds, which can degrade metric stability in absence of sufficient positive samples.
These results reinforce the robustness of ImLPR under varying spatial tolerances, confirming its capacity for precise place recognition.

\section{Robustness to Yaw Variation}

\subsection{Place Recognition Performance under Yaw Variation}

\begin{figure}[!t]
    \centering
    \begin{minipage}{0.7\textwidth}  
        \centering
        \includegraphics[trim = 0cm 0cm 0cm 0cm, width=\linewidth]{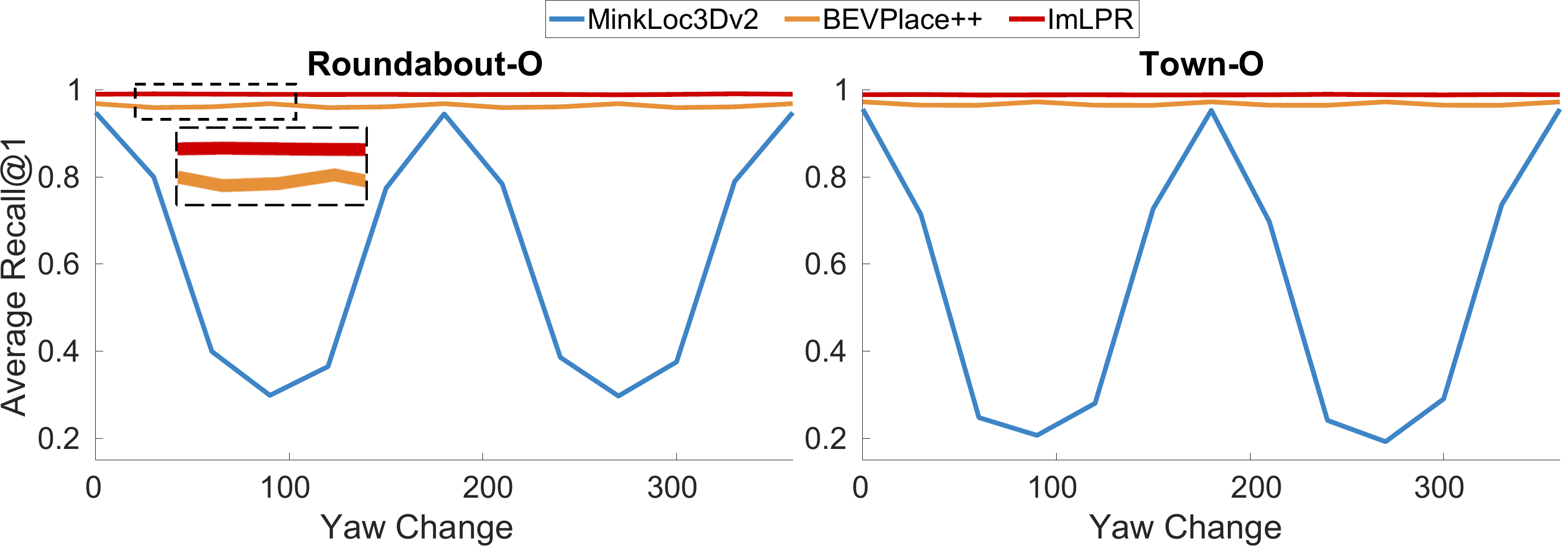}
        \vspace{-5mm}
    \end{minipage}%
    \hfill
    \begin{minipage}{0.3\textwidth}  
        \centering
        \begin{table}[H]
            \caption{$\sigma$ of AR@1}
            \centering
            \resizebox{\linewidth}{!}{
                \begin{tabular}{l|c|c}
                    \toprule
                    {Method} & \texttt{Round-O} & \texttt{Town-O} \\
                    \midrule
                    MinkLoc3Dv2 & 26.900 & 31.347 \\
                    BEVPlace++ & \underline{0.420} & \underline{0.389} \\
                    ImLPR & \textbf{0.064} & \textbf{0.047} \\
                    \bottomrule
                \end{tabular}
            }
                \vspace{8mm}           
            \label{tab:recall_std_yaw}
        \end{table}
    \end{minipage}
    \caption{Average Recall@1 and its standard deviation ($\sigma$) for inter-session place recognition across yaw variations. ImLPR exhibits inherent yaw robustness, achieving the lowest $\sigma$.}
    \vspace{-6mm}
    \label{fig:yaw_invariant}
\end{figure}

Descriptors for identical locations should maintain consistency despite sensor rotations, particularly yaw angle variations, which frequently occur during scene revisit. To assess this, we applied yaw rotations to database scans from the \texttt{Roundabout-O} and \texttt{Town-O} sequences and compared them against unrotated query scans. The Average Recall@1, obtained from inter-session place recognition experiments across all evaluated sequences, is depicted in \figref{fig:yaw_invariant}.

As illustrated in \figref{fig:yaw_invariant}, MinkLoc3Dv2 exhibits performance degradation with varying yaw angles, reflecting limited robustness to such transformations. BEVPlace++ incorporates a rotation-equivariant module to mitigate yaw variance. This module rotates the input image at a fixed angular interval of $\theta = 45^\circ$, processes each rotated image through a convolutional neural network, and applies max pooling across the resultant features. While this approach enhances the yaw variation robustness, performance fluctuates slightly when the variations deviate from the predefined discrete angles used in max pooling. In contrast, ImLPR generates yaw-robust descriptors without additional computational overhead, owing to its architectural design.

For a RIV image, yaw variation is equivalent to a horizontal shift. DINOv2, the vision transformer we use, processes patch features via attention mechanisms that capture global inter-patch relationships. Since horizontal shifts reorder patches without modifying their content, the attention mechanism yields nearly translation-equivariant feature representations, preserving the relational structure. Consequently, vision transformers produce similar patch features that are shifted from the original patch features, alongside consistent global tokens for both original and shifted images. Similarly, the MultiConv adapter employs 2D convolutions, which are inherently translation-equivariant, ensuring that a horizontal shift in the input image results in a corresponding shift in patch features while maintaining their values. These refined patch features are aggregated using SALAD’s optimal transport approach, which is invariant to horizontal shifts, as demonstrated in \secref{sec:proof_permutation}. Data augmentation strategies, including masking and random transformations, further bolster this stability. However, positional encoding introduces a minor constraint: as yaw rotation shifts the image, identical patches receive different positional encodings, leading to subtle differences in the vision transformer’s output before and after rotation. Despite this, ImLPR achieves more consistent performance than BEVPlace++ under yaw variation, as shown in \tabref{tab:recall_std_yaw}. This highlights the robustness of ImLPR’s model architecture and RIV representation in handling yaw variations effectively.

\subsection{Horizontal Shift Invariance in SALAD’s Optimal Transport Aggregation}
\label{sec:proof_permutation}
In this section, we prove that SALAD’s optimal transport (OT) aggregation generates an identical global descriptor despite horizontal (column-wise) shifts in the input feature map. Let \(\mathbf{F} \in \mathbb{R}^{H'W' \times C}\) be the flattened local features derived from adapter-refined patch features, originating from an un-flattened feature map \(\mathbf{F}' \in \mathbb{R}^{H' \times W' \times C}\) with \(n = H'W'\) patches. A translation-equivariant convolutional layer maps \(\mathbf{F}'\) to a score map \(\mathbf{S}' \in \mathbb{R}^{H' \times W' \times m}\), which is flattened to \(\mathbf{S} \in \mathbb{R}^{n \times m}\), representing assignment probabilities of \(n\) patches to \(m\) learnable cluster centers. A column shift in \(\mathbf{F}'\) (e.g., \(\mathbf{F}'_{p,q,c} \to \mathbf{F}'_{p,\mathrm{mod}((q-s), W'),c}\)) induces an identical column shift in \(\mathbf{S}'\): \(\mathbf{S}'_{p,q,m} \to \mathbf{S}'_{p,\mathrm{mod}((q-s), W'),m}\). Here, \(p \in [1, H']\) and \(q \in [1, W']\) denote the row and column indices, and the function \(\mathrm{mod}(a, b) = ((a-1) \mod b) + 1\) ensures the result lies in \([1, b]\). After flattening, this translates to a row shift in \(\mathbf{S}\): \(\mathbf{S}_{i,:} \to \mathbf{S}_{\mathrm{mod}((i-s'), n),:}\), where \(s' = s \times H'\). This preserves the score values for each patch, only reordering their row indices, ensuring shift invariance in the subsequent OT aggregation.

For OT aggregation, the Sinkhorn algorithm~\cite{NIPS2013_af21d0c9} solves an entropically regularized OT problem. The score matrix \(\mathbf{S} \in \mathbb{R}^{n \times m}\) represents assignment probabilities to \(m\) cluster centers. The cost matrix is defined as \(\mathbf{M} = -\mathbf{S} / \lambda\), where \(\lambda\) is the regularization parameter controlling the entropy of the transport plan. A column shift in \(\mathbf{S}\) results in a row shift in \(\mathbf{M}\). The Sinkhorn algorithm iteratively updates dual variables \(\mathbf{u} \in \mathbb{R}^{n}\) and \(\mathbf{v} \in \mathbb{R}^{m}\) to satisfy marginal constraints, using uniform log-weights \(\log \mathbf{a}\) and \(\log \mathbf{b}\). Since \(\log \mathbf{a}\) is uniform across patches, \(\mathbf{u}\) shifts consistently under a row shift. Specifically, for patch \(i\):
\begin{equation}
u_i = \log a_i - \log \sum_{k=1}^{m} \exp(M_{i,k} + v_k).
\label{eq:u_update}
\end{equation}
A row shift (e.g., \(\mathbf{M}_{i,k} \to \mathbf{M}_{\mathrm{mod}((i-s'), n),k}\)) results in \(u_i \to u_{\mathrm{mod}((i-s'), n)}\). The OT matrix is computed as:
\begin{equation}
\log R_{i,k} = M_{i,k} + u_i + v_k.
\label{eq:ot_matrix}
\end{equation}
Consequently, the rows of \(\log \mathbf{R}\) shift identically to \(\mathbf{S}\), preserving assignment weights, yielding \(\mathbf{R} \in \mathbb{R}^{n \times m}\). The aggregated cluster features are computed as a weighted sum of intermediate feature embeddings \(\bar{\mathbf{F}} \in \mathbb{R}^{n \times l}\), derived by applying a convolutional layer to \(\mathbf{F}'\) and flattened from \(\bar{\mathbf{F}}' \in \mathbb{R}^{H' \times W' \times l}\):
\begin{equation}
V_{j,k} = \sum_{i=1}^{n} R_{i,j} \bar{F}_{i,k}, \quad j = 1, \dots, m, \quad k = 1, \dots, l,
\label{eq:cluster_agg}
\end{equation}
producing \(\mathbf{V} \in \mathbb{R}^{m \times l}\). The feature matrix \(\bar{\mathbf{F}}\) inherits the same column shift: \(\bar{\mathbf{F}}'_{p,q,l} \to \bar{\mathbf{F}}'_{p,\mathrm{mod}((q-s), W'),l}\), or \(\bar{\mathbf{F}}_{i,:} \to \bar{\mathbf{F}}_{\mathrm{mod}((i-s'), n),:}\) after flattening. Since \(\bar{\mathbf{F}}_{i,k}\) and \(R_{i,j}\) shift identically, and summation is commutative, the aggregated features are invariant to column shifts:
\begin{equation}
\sum_{i=1}^{n} R_{\mathrm{mod}((i-s'), n),j} \bar{F}_{\mathrm{mod}((i-s'), n),k} = \sum_{i=1}^{n} R_{i,j} \bar{F}_{i,k}.
\label{eq:shift_inv}
\end{equation}
The global descriptor concatenates the flattened cluster features \(\mathbf{V} \in \mathbb{R}^{m \times l}\) with the global embedding \(\mathbf{G} \in \mathbb{R}^{e}\), processed independently via linear layers and unaffected by shifts, forming \(g = [\mathbf{V}.\text{flatten}(), \mathbf{G}] \in \mathbb{R}^{m \times l + e}\), ensuring horizontal shift invariance.

\section{Ablation Studies}
To assess the impact of individual components in ImLPR for place recognition, we perform a series of ablation studies. We evaluate performance using AR@1 and AF1 for inter-session place recognition. These studies systematically analyze the contributions of key elements to ImLPR’s overall effectiveness. The results provide insights into the significance of each component in achieving robust and efficient \ac{LPR} performance.

\input{tab/ab_adapter}
\subsection{MultiConv Adapter and the Number of Trained Block in DINOv2}
\ac{VFM}s, such as DINOv2, leverage extensive pre-training on large-scale datasets to deliver robust feature representations. To preserve these learned representations during adaptation for \ac{LPR}, it is essential to avoid catastrophic forgetting while effectively bridging the domain gap between natural vision images and RIV images. Fine-tuning multiple transformer blocks risks overwriting the model’s general knowledge, whereas minimal fine-tuning with strategic adaptations can maintain performance. Following the main evaluation, this section examines how the inclusion of the MultiConv adapter and varying the number of trained transformer blocks influence the performance of ImLPR.

The results, presented in \tabref{tab:ab_adapter}, highlight the critical role of the MultiConv adapter. Without the adapter and without fine-tuning (\texttt{ExpC-1}), the model fails to perform effective place recognition due to the significant domain gap between natural images and RIV images. Fine-tuning without the adapter (\texttt{ExpC-2}) also results in reduced performance, as the limited number of trainable parameters hinders the model’s ability to extract robust features from RIV images. In contrast, incorporating the MultiConv adapter while keeping most transformer blocks frozen (\texttt{ExpC-3} and \texttt{ExpC-4}) significantly improves performance. This demonstrates the adapter’s ability to efficiently address domain shifts, leveraging DINOv2’s pre-trained knowledge for enhanced place recognition.

Additionally, we analyze the effect of varying the number of trained transformer blocks alongside the MultiConv adapter. For \texttt{Roundabout-O} and \texttt{Town-O}, performance remains stable across configurations, indicating that effective domain adaptation can be achieved with the adapter and minimal fine-tuning of just a few transformer blocks. However, for \texttt{DCC} and \texttt{Roundabout-V}, performance declines as more blocks are trained. This degradation likely results from overfitting to the training dataset due to an increased number of trainable parameters, which compromises the model’s ability to generalize. These findings highlight the critical role of preserving pre-trained knowledge and the necessity of the MultiConv adapter in effectively balancing domain adaptation, ensuring robust \ac{LPR} performance with minimal fine-tuning.

\input{tab/ab_augmentation}

\subsection{Effect of Data Augmentation}

We evaluate the impact of data augmentation on ImLPR’s inter-session place recognition performance, with results presented in \tabref{tab:ab_augmentation}. Without any augmentations (\texttt{ExpD-1}), the model achieves baseline performance but struggles with orientation variability and projection artifacts in RIV images, limiting its robustness. Applying random yaw rotation (\texttt{ExpD-2}) significantly improves performance by introducing column shifts that enhance the model’s resilience to orientation changes in LiDAR scans. Further improvement is observed with the addition of line and square masking (\texttt{ExpD-3}). Both masks simulate sparse regions by introducing empty pixels in RIV images, mimicking the effect of dropout and training the model to perform place recognition with partially empty patch, thus boosting robustness to incomplete inputs.

The best performance is achieved when combining all augmentations—random yaw rotation, line, square, and cylindrical masking (\texttt{ExpD-4}). The cylindrical mask also enhances the model’s ability to handle the extreme occlusions by training it to rely on partial RIV images. This configuration delivers superior results across all datasets, with notable gains on \texttt{DCC} and \texttt{Roundabout-V}, which include occlusions in their scans. These results underscore the vital role of data augmentation in enhancing ImLPR’s robustness, with the combined yaw rotation and masking strategies effectively addressing orientation variability, projection artifacts, and scene sparsity in \ac{LPR}.

\subsection{Dimension of DINOv2}

The impact of varying the DINOv2 backbone’s feature dimension on ImLPR’s inter-session place recognition performance and computational efficiency is assessed in \tabref{tab:ab_dim}. All experiments here use a single NVIDIA GeForce RTX 3090 GPU to measure computation time. This ablation study evaluates three DINOv2 configurations—ViT-S/14, ViT-B/14, and ViT-L/14—with feature dimensions of 384, 768, and 1024, respectively, analyzing their place recognition efficacy and computational cost. The final two transformer blocks are fine-tuned for all models to ensure consistent adaptation, and computational efficiency is compared against other \ac{SOTA} methods.

\subsubsection{Performance Analysis}

The place recognition performance of ImLPR across different feature dimensions is detailed in \tabref{tab:ab_dim}, which presents results for multiple datasets. All three configurations—ViT-S/14 (\texttt{ExpE-1}), ViT-B/14 (\texttt{ExpE-2}), and ViT-L/14 (\texttt{ExpE-3})—exhibit similar performance, achieving robust place recognition outcomes. This similarity indicates that higher feature dimensions do not always yield proportional improvements in accuracy. Notably, the smallest model, ViT-S/14, demonstrates exceptional robustness, particularly on \texttt{Roundabout-O}, \texttt{Town-O}, and \texttt{Roundabout-V}, where it delivers highly competitive results. These findings indicate that larger models may not always outperform smaller ones, as their increased complexity can hinder effective training on diverse RIV images and lead to overfitting on the training data.

\input{tab/ab_dim_time}

\subsubsection{Computational Cost}

The computational efficiency is critical for real-world place recognition applications. As reported in \tabref{tab:ab_dim}, the time required for descriptor extraction scales with feature dimension: ViT-S/14 (\texttt{ExpE-1}) is the most efficient, followed by ViT-B/14 (\texttt{ExpE-2}), with ViT-L/14 (\texttt{ExpE-3}) incurring the highest computational cost. All configurations achieve extraction times below 100ms, demonstrating their suitability for real-time performance in \ac{LPR}. As depicted in \figref{fig:ab_dim}, we further compare ImLPR against baseline methods using Average Recall@1 on \texttt{Roundabout-O} and \texttt{Town-O}. BEVPlace++ requires 27.5ms for descriptor extraction, despite a smaller feature dimension of 128, surpassing the runtime of ViT-S/14 and ViT-B/14, due to its reliance on multiple ResNet instances for yaw invariance. MinkLoc3Dv2, utilizing 3D sparse convolution, achieves the fastest extraction time but yields the lowest Average Recall@1. In contrast, CASSPR’s attention-based networks and per-point feature extraction lead to the longest computation time. ImLPR’s Vision Transformer-based backbone inherently supports robustness to yaw variation and achieve a low latency of 18.1ms. Furthermore, as shown in \figref{fig:ab_dim}, ImLPR achieves the highest Average Recall@1, demonstrating its suitability for real-time place recognition applications requiring minimal latency.

The balance between performance and computational cost is a critical factor in selecting an optimal backbone. The results suggest that overly large models, such as ViT-L/14, may introduce unnecessary complexity without corresponding performance benefits, potentially due to overfitting or challenges in training. Conversely, ViT-S/14 offers an optimal trade-off between model size, computational efficiency, and the performance of place recognition, making it highly suitable for practical deployment.

\begin{table}[t!]
\centering
\caption{Average Results of Intra-session PR with Visual Place Recognition models}
\label{tab:VPR}
\resizebox{0.90\columnwidth}{!}{%
\begin{tabular}{l|c|c||cc|cc|cc} \toprule
  & DINOv2 & {Runtime}                & \multicolumn{2}{c|}{\texttt{Roundabout-O}}   & \multicolumn{2}{c|}{\texttt{Town-O}}             & \multicolumn{2}{c}{Average}                \\ 
Method & Backbone & {(ms)} & \multicolumn{1}{c}{AR@1} & \multicolumn{1}{c|}{AF1} & \multicolumn{1}{c}{AR@1} & \multicolumn{1}{c|}{AF1} & \multicolumn{1}{c}{AR@1} & \multicolumn{1}{c}{AF1} \\ \midrule
SALAD & ViT-B/14 & 21.3  & 0.841 & 0.919  & 0.771 & 0.874 & 0.806 & 0.896 \\
BoQ   & ViT-B/14 & 22.7   & 0.934 & 0.967 & 0.896 & 0.947 & 0.915 & 0.957\\
SelaVPR (global) & ViT-L/14& 79.5 & 0.955 &  0.978 & 0.941 & 0.970 & 0.948 &  0.974\\
ImLPR & ViT-S/14 &\textbf{18.1}  & \textbf{0.986} & \textbf{0.994} & \textbf{0.966} & \textbf{0.984} & \textbf{0.976} & \textbf{0.989} \\ \bottomrule
\end{tabular}%
}
\vspace{-2.5mm}
\end{table}

\section{Comparison with Visual Place Recognition Models}

ImLPR employs a \ac{VFM} and adopts image-based aggregation strategies, SALAD, similar to those used in \ac{VPR}. To assess whether RIV images—composed of reflectivity, range, and normal ratio channels—can be effectively processed by conventional VPR models, we compare ImLPR with recent VPR methods, including SALAD \cite{izquierdo2024optimal}, BoQ \cite{ali2024boq}, and SelaVPR \cite{lu2024towards}. To ensure fair comparison, all models were trained using the same RIV inputs and identical data augmentation strategies. We also used the same loss function (TSAP loss) across all methods, and all feature encoders and training hyperparameters were kept at their default settings.

\tabref{tab:VPR} reports average intra-session place recognition performance on \texttt{Roundabout-O} and \texttt{Town-O}. While existing VPR methods achieve high accuracy in traditional visual tasks, they show limited performance when directly applied to RIV images. SALAD and BoQ, which use ViT-B/14 backbones trained with the TSAP loss, underperform compared to ImLPR across all metrics, indicating that naive application of VPR pipelines to LiDAR-based images does not adequately handle the sensor domain gap. SelaVPR, which integrates adapters and uses a larger ViT-L/14 backbone, performs better than SALAD and BoQ but still falls short of ImLPR.

Despite using a smaller backbone (ViT-S/14), ImLPR outperforms all baselines by a notable margin and operates with the lowest runtime (18.1 ms). This highlights both the computational efficiency and accuracy of the proposed method. The performance gap emphasizes the importance of explicitly addressing the domain shift between camera and LiDAR data. In particular, the use of adapters allows for better alignment with LiDAR-specific structures, as evidenced in both SelaVPR and ImLPR, while our Patch-InfoNCE loss further enhances RIV feature learning by leveraging range-aware supervision, which is unavailable in image-based models. These results demonstrate that high performance on LiDAR images cannot be achieved by merely applying VPR models composed of DINOv2 and an aggregation network to RIV inputs. Instead, effective place recognition requires proper adaptation beyond simple fine-tuning, as well as LiDAR-aware feature learning.

\color{black}

\section{Comparison with 3D Foundation Model}
\label{sec:3d_foundation}
In this section, we evaluate ImLPR against a 3D foundation model, specifically PointTransformerv3 (PTv3). Other 3D foundation models, such as Sonata, could potentially serve as feature extractors; however, Sonata lacks a pre-trained model trained on outdoor datasets and requires color information, which is incompatible with raw LiDAR scans. To align PTv3 with outdoor \ac{LPR}, we re-train its pre-trained model, originally trained on the nuScenes dataset\footnotemark[1]\footnotetext[1]{https://www.nuscenes.org/}, with the HeLiPR-O dataset. This re-training uses $x, y, z$ coordinates and reflectivity from the point cloud. For consistency with ImLPR, we aggregate PTv3’s output features using the SALAD. To ensure computational efficiency, we turn on the flash attention for PTv3 and both methods are tested on a single NVIDIA GeForce RTX 4090 GPU. Additionally, we apply the same augmentation strategy used during the original training of PTv3 on the nuScenes dataset.

\input{tab/ab_3d_foundation}

As shown in \tabref{tab:3d_foundation}, ImLPR is more lightweight, with approximately half the parameters of PTv3 and a runtime four times faster. Although PTv3 surpasses baselines like CASSPR and MinkLoc3Dv2 in \tabref{tab:intra}, \tabref{tab:inter_roundabout}, and \tabref{tab:inter_town} thanks to its ability of 3D foundation model, ImLPR consistently outperforms PTv3 in both intra-session and inter-session \ac{LPR} while maintaining superior computational efficiency. These results demonstrate that ImLPR’s adaptation of LiDAR data to the vision domain delivers a more effective and efficient solution for \ac{LPR}. By leveraging DINOv2’s extensive pre-trained knowledge from millions of images, ImLPR outperforms PTv3, which is trained on thousands of point clouds. This underscores the advantage of a \ac{VFM} with a well-crafted approach, enabling it to surpass 3D foundation models even within the LiDAR domain for \ac{LPR}.

\begin{figure}[!t]
    \centering
    \includegraphics[trim = 0cm 0cm 0cm 0cm, width=\columnwidth]{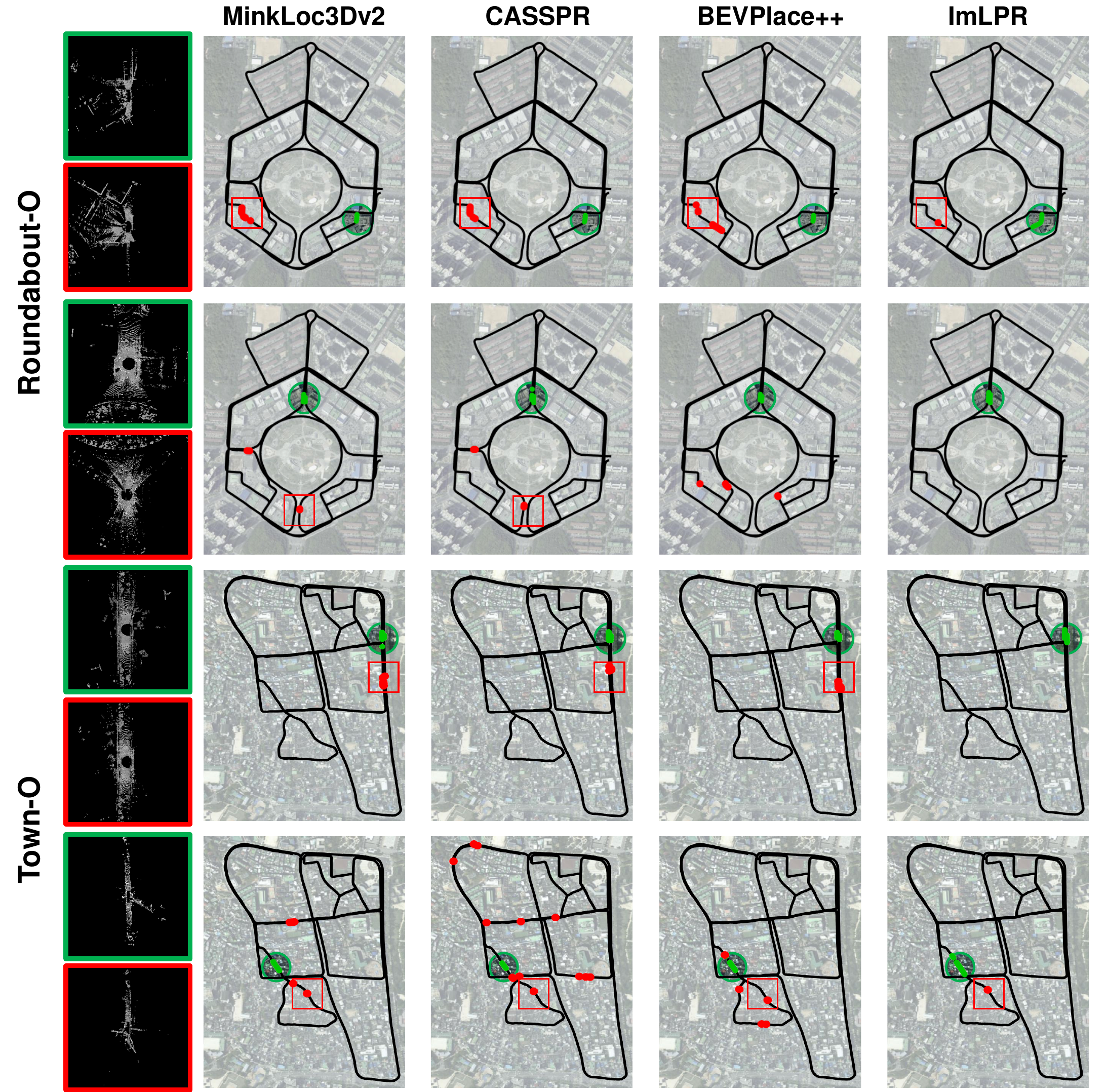}
    \caption{We retrieved 20 locations from \texttt{Roundabout01-O} and \texttt{Town01-O} using queries from \texttt{Roundabout02-O} and \texttt{Town02-O}. Each image, overlaid with trajectory and satellite imagery, indicates the query with a green circle, true positives with green points, and false positives with red points. Additionally, BEV images illustrate the sensing environments, marking the query location in green and incorrect retrieved locations with a red square.}
    \label{fig:qualitative}
    \vspace{-2mm}
\end{figure}

\section{Additional Qualitative Results}

In this section, we analyze the retrieval distribution for inter-session place recognition, defining a positive match as within 50 meters, using \texttt{Roundabout01-O} and \texttt{Town01-O} as the database and queries from \texttt{Roundabout02-O} and \texttt{Town02-O}. As illustrated in \figref{fig:qualitative}, ImLPR retrieves nearly all candidates closer to the query compared to other methods, indicating that its descriptors preserve small feature distances for both the top-1 match and proximate locations. This demonstrates that RIV images effectively project geometric space into the descriptor embedding space. In contrast, other approaches generate multiple false positives at certain locations due to strong scene appearance similarity with the query, leading to closely related descriptors and an inability to distinguish places. For example, in the upper case of \texttt{Town01-O}, BEV images exhibit highly similar scene contours, highlighting the challenge of differentiating locations based solely on geometric structure. This emphasizes the value of multi-channel strategies like ImLPR, which enhance \ac{LPR} by incorporating diverse inputs beyond mere geometry.

%% file: tab/gen_dcc.tex
\begin{table}[t]
\centering
\caption{Performance of High-Resolution LiDAR-Trained Models on Low-Resolution LiDAR}
\label{tab:sensor_comparison}
\resizebox{\textwidth}{!}{%
\begin{tabular}{l||cc|cc|cc|cc||cc|cc|cc|cc||cc}
\toprule
 &  
\multicolumn{8}{c||}{NCLT (HDL-32E)\texttt{}} & 
\multicolumn{8}{c||}{HeLiPR \texttt{Roundabout-V} (VLP-16C)} & 
\multicolumn{2}{c}{Average} \\ 
\cline{2-9} \cline{10-17} 
\multirow{1}{*}{Method} & 
\multicolumn{2}{c|}{\texttt{2012-01-08}} & 
\multicolumn{2}{c|}{\texttt{2012-01-15}} & 
\multicolumn{2}{c|}{\texttt{2012-01-22}} & 
\multicolumn{2}{c||}{Average} & 
\multicolumn{2}{c|}{\texttt{01}} & 
\multicolumn{2}{c|}{\texttt{02}} & 
\multicolumn{2}{c|}{\texttt{03}} & 
\multicolumn{2}{c||}{Average} & 
\multicolumn{2}{c}{\texttt{}} \\ 
 & R@1 & F1 & R@1 & F1 & R@1 & F1 & R@1 & F1 & R@1 & F1 & R@1 & F1 & R@1 & F1 & R@1 & F1 & R@1 & F1 \\ 
\midrule
LoGG3D-Net & 
  0.173 & 0.568 & 0.141 & 0.416 & 0.132 & 0.318 & 0.149 & 0.434 & 0.038 & 0.358 & 0.052 & 0.133 & 0.302 & 0.690 & 0.131 & 0.394 & 0.140 & 0.414 \\
MinkLoc3dv2 & 
  0.649 & 0.836 & 0.603 & 0.773 & 0.613 & 0.815 & 0.622 & 0.808 & 0.469 & 0.771 & 0.421 & 0.634 & 0.648 & \underline{0.887} & 0.513 & 0.764 & 0.567 & 0.786 \\
CASSPR & 
  0.681 & 0.839 & 0.623 & 0.789 & 0.651 & 0.801 & 0.652 & 0.810 & 0.607 & \underline{0.784} & 0.571 & 0.728 & 0.711 & 0.867 & 0.630 & \underline{0.793} & 0.641 & 0.801 \\
BEVPlace++ & 
  \textbf{0.861} & \textbf{0.928} & \textbf{0.838} & \underline{0.915} & \underline{0.850} & \underline{0.923} & \textbf{0.850} & \underline{0.922} & \underline{0.624} & 0.769 & \underline{0.575} & \underline{0.731} & \textbf{0.767} & 0.870 & \underline{0.655} & 0.790 & \underline{0.753} & \underline{0.856} \\
ImLPR & 
  \underline{0.817} & \underline{0.920} & \underline{0.830} & \textbf{0.930} & \textbf{0.854} & \textbf{0.932} & \underline{0.834} & \textbf{0.927} & \textbf{0.765} & \textbf{0.884} & \textbf{0.617} & \textbf{0.781} & \underline{0.753} & \textbf{0.892} & \textbf{0.712} & \textbf{0.852} & \textbf{0.773} & \textbf{0.890} \\
\bottomrule
\end{tabular}%
}
\vspace{-3mm}
\end{table}

%% file: tab/ab_adapter.tex
\begin{table}[t]
\centering
\caption{Ablation study for adapter and trained block}
\label{tab:ab_adapter}
\resizebox{\textwidth}{!}{%
\begin{tabular}{l||cc||cc|cc|cc|cc|cc}
\toprule
\multirow{2}{*}{Method} &
  \multirow{2}{*}{Adapter} &
  \multirow{2}{*}{\begin{tabular}[c]{@{}c@{}}Trained\\ Block\end{tabular}} &
  \multicolumn{2}{c|}{\texttt{Roundabout-O}} &
  \multicolumn{2}{c|}{\texttt{Town-O}} &
  \multicolumn{2}{c|}{\texttt{DCC}} &
  \multicolumn{2}{c|}{\texttt{Roundabout-V}} &
  \multicolumn{2}{c}{Average} \\
 & & &
  AR@1 & AF1 &
  AR@1 & AF1 &
  AR@1 & AF1 &
  AR@1 & AF1 &
  AR@1 & AF1 \\
\midrule
\texttt{ExpC-1} & & 0 & 0.321 & 0.530 & 0.379 & 0.553 & 0.376 & 0.555 & 0.249 & 0.464 & 0.331 & 0.526 \\
\texttt{ExpC-2} & & 2 & 0.850 & 0.925 & 0.833 & 0.913 & 0.850 & 0.943 & 0.645 & 0.809 & 0.795 & 0.898 \\
\texttt{ExpC-3} & \ding{52} & 0 & 0.972 & 0.987 & 0.965 & 0.983 & 0.928 & 0.965 & 0.799 & 0.906 & 0.916 & 0.960 \\
\texttt{ExpC-4} & \ding{52} & 2 & \textbf{0.990} & \textbf{0.996} & \underline{0.989} & \underline{0.995} & \textbf{0.942} & \textbf{0.973} & \textbf{0.888} & \textbf{0.948} & \textbf{0.952} & \textbf{0.978} \\
\texttt{ExpC-5} & \ding{52} & 6 & \textbf{0.990} & \underline{0.995} & \textbf{0.990} & \textbf{0.996} & \underline{0.941} & \textbf{0.973} & 0.868 & \underline{0.940} & \underline{0.947} & \underline{0.976} \\
\texttt{ExpC-6} & \ding{52} & 10 & \underline{0.986} & 0.994 & 0.988 & \underline{0.995} & 0.935 & \underline{0.972} & \underline{0.870} & 0.936 & 0.945 & 0.974 \\
\bottomrule
\end{tabular}%
}
\vspace{-4mm}
\end{table}

%% file: tab/ab_augmentation.tex
\begin{table}[t]
\centering
\caption{Ablation study for yaw change and mask types}
\label{tab:ab_augmentation}
\resizebox{\columnwidth}{!}{%
\begin{tabular}{l||ccc||cc|cc|cc|cc|cc}
\toprule
\multirow{2}{*}{Method} &
  \multirow{2}{*}{\begin{tabular}[c]{@{}c@{}}Yaw\\ Change\end{tabular}} &
  \multirow{2}{*}{\begin{tabular}[c]{@{}c@{}}Line \& Square\\ Mask\end{tabular}} &
  \multirow{2}{*}{\begin{tabular}[c]{@{}c@{}}Cylindrical\\ Mask\end{tabular}} &
  \multicolumn{2}{c|}{\texttt{Roundabout-O}} &
  \multicolumn{2}{c|}{\texttt{Town-O}} &
  \multicolumn{2}{c|}{\texttt{DCC}} &
  \multicolumn{2}{c|}{\texttt{Roundabout-V}} &
  \multicolumn{2}{c}{Average} \\
 & & & &
  AR@1 & AF1 &
  AR@1 & AF1 &
  AR@1 & AF1 &
  AR@1 & AF1 &
  AR@1 & AF1 \\
\midrule
\texttt{ExpD-1} & & & &
  0.937 & 0.972 &
  0.943 & 0.973 &
  \underline{0.905} & 0.966 &
  0.781 & 0.893 &
  0.892 & 0.951 \\
\texttt{ExpD-2} & \ding{52} & & &
  0.980 & 0.990 &
  0.958 & 0.980 &
  0.896 & 0.983 &
  0.821 & 0.927 &
  0.914 & 0.970 \\
\texttt{ExpD-3} & \ding{52} & \ding{52} & &
  \underline{0.984} & \underline{0.992} &
  \underline{0.975} & \underline{0.989} &
  0.894 & \underline{0.985} &
  \underline{0.837} & \underline{0.928} &
  \underline{0.923} & \underline{0.974} \\
\texttt{ExpD-4} & \ding{52} & \ding{52} & \ding{52} &
  \textbf{0.990} & \textbf{0.996} &
  \textbf{0.989} & \textbf{0.995} &
  \textbf{0.942} & \textbf{0.973} &
  \textbf{0.888} & \textbf{0.948} &
  \textbf{0.952} & \textbf{0.978} \\
\bottomrule
\end{tabular}%
}
\vspace{-5mm}
\end{table}

%% file: tab/ab_dim_time.tex
\begin{figure}[!t]
    \centering
\begin{minipage}{0.74\textwidth}  
        \centering
        \begin{table}[H]
        \centering
        \caption{Ablation study for feature dimension and computation time}
        \label{tab:ab_dim}
        \resizebox{\columnwidth}{!}{%
        \begin{tabular}{l||cc||cc|cc|cc|cc|cc}
        \toprule
        \multirow{2}{*}{Method} &
          \multirow{2}{*}{\begin{tabular}[c]{@{}c@{}}Feature\\ Dim.\end{tabular}} &
          \multirow{2}{*}{\begin{tabular}[c]{@{}c@{}}Runtime\\ (ms)\end{tabular}} &
          \multicolumn{2}{c|}{\texttt{Roundabout-O}} &
          \multicolumn{2}{c|}{\texttt{Town-O}} &
          \multicolumn{2}{c|}{\texttt{DCC}} &
          \multicolumn{2}{c|}{\texttt{Roundabout-V}} &
          \multicolumn{2}{c}{Average} \\
         & & &
          AR@1 & AF1 &
          AR@1 & AF1 &
          AR@1 & AF1 &
          AR@1 & AF1 &
          AR@1 & AF1 \\
        \midrule
        \texttt{ExpE-1} & 384 & 18.1 &
          \textbf{0.990} & \textbf{0.996} &
          \textbf{0.989} & \textbf{0.995} &
          0.942 & \underline{0.973} &
          \textbf{0.888} & \textbf{0.948} &
          \textbf{0.952} & \textbf{0.978} \\
        \texttt{ExpE-2} & 768 & 23.3 &
          \underline{0.989} & \underline{0.995} &
          \underline{0.988} & \underline{0.994} &
          \underline{0.943} & \underline{0.973} &
          0.861 & 0.942 &
          0.945 & \underline{0.976} \\
        \texttt{ExpE-3} & 1024 & 58.0 &
          \underline{0.989} & \underline{0.995} &
          0.986 & \underline{0.994} &
          \textbf{0.957} & \textbf{0.979} &
          \underline{0.873} & \underline{0.944} &
          \underline{0.951} & \textbf{0.978} \\
        \bottomrule
        \end{tabular}%
        }
        \end{table}
        \vspace{5mm}
    \end{minipage}
    \hfill
    \begin{minipage}{0.25\textwidth}  
        \centering
        \includegraphics[trim = 0cm 0cm 0cm 0cm, width=\linewidth]{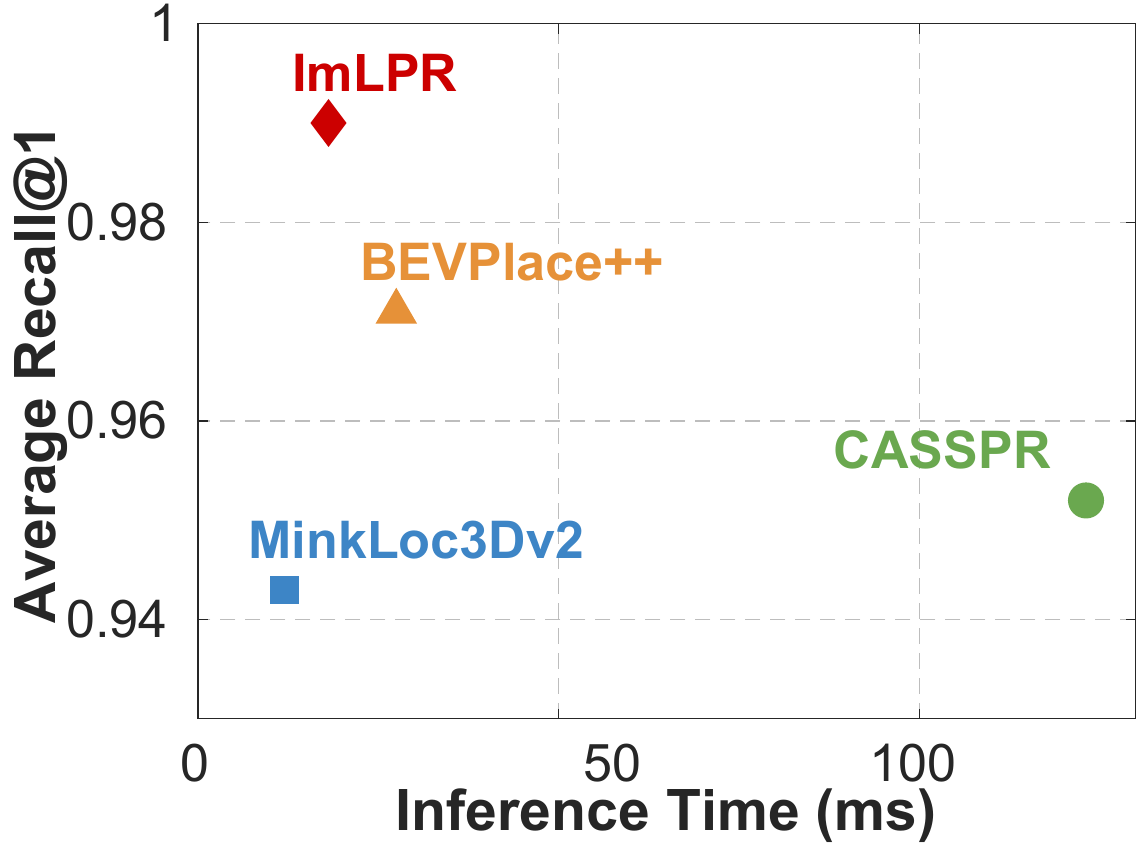}
        \vspace{-5mm}
        \
    \end{minipage}%
    
    \vspace{-3mm}
    \caption{(Left) \tabref{tab:ab_dim} displays feature dimensions and inference times for ViT models, showing comparable performance despite larger dimensions. (Right) The plot illustrates inference time versus Average Recall@1 on \texttt{Roundabout-O} and \texttt{Town-O}, with ImLPR achieving the highest performance and an inference time comparable to MinkLoc3Dv2 for descriptors.}
    \vspace{-5mm}
    \label{fig:ab_dim}
\end{figure}

%% file: tab/ab_3d_foundation.tex
\begin{table}[!t]
\centering
\caption{Performance Comparison of 2D and 3D Foundation Models}
\label{tab:3d_foundation}
\resizebox{\columnwidth}{!}{%
\begin{tabular}{l||cc||cc|cc|cc|cc}
\toprule
\multirow{3}{*}{Method} &
  \multirow{3}{*}{\begin{tabular}[c]{@{}c@{}}Params\\ (M)\end{tabular}} &
  \multirow{3}{*}{\begin{tabular}[c]{@{}c@{}}Runtime\\ (ms)\end{tabular}} &
  \multicolumn{4}{c|}{Intra-session LPR} &
  \multicolumn{4}{c}{Inter-session LPR} \\ \cline{4-11}
 & & &
  \multicolumn{2}{c|}{\texttt{Roundabout-O}} &
  \multicolumn{2}{c|}{\texttt{Town-O}} &
  \multicolumn{2}{c|}{\texttt{Roundabout-O}} &
  \multicolumn{2}{c}{\texttt{Town-O}} \\
 & & &
  AR@1 & AF1 &
  AR@1 & AF1 &
  AR@1 & AF1 &
  AR@1 & AF1 \\
\midrule
{PTv3+SALAD} & 46.3 & 55.0 &
  0.927 & 0.964 &
  0.938 & 0.969 &
  0.969 & 0.985 &
  0.970 & 0.985 \\
{ImLPR} & \textbf{25.7} & \textbf{12.9} &
  \textbf{0.986} & \textbf{0.994} &
  \textbf{0.966} & \textbf{0.984} &
  \textbf{0.990} & \textbf{0.996} &
  \textbf{0.989} & \textbf{0.995} \\
\bottomrule
\end{tabular}%
}
\vspace{-4mm}
\end{table}